\documentclass[aip,
 amsmath,amssymb,
 reprint,%
]{revtex4-1}
\usepackage{amsmath}
\usepackage{amssymb}
\usepackage{graphicx}
\usepackage{dcolumn}
\usepackage{bm}
\usepackage[hidelinks,colorlinks=true,linkcolor=blue,citecolor=blue]{hyperref}

\usepackage[utf8]{inputenc}
\usepackage[T1]{fontenc}
\usepackage{mathptmx}
\usepackage{etoolbox}
\usepackage{mathrsfs}
\usepackage{subcaption}
\usepackage{xcolor}
\usepackage[english]{babel}
\usepackage[mathscr]{euscript}

\makeatletter
\def\@email#1#2{%
 \endgroup
 \patchcmd{\titleblock@produce}
  {\frontmatter@RRAPformat}
  {\frontmatter@RRAPformat{\produce@RRAP{*#1\href{mailto:#2}{#2}}}\frontmatter@RRAPformat}
  {}{}
}%
\makeatother
\begin{document}


\title[]{Transfer learning-enhanced deep reinforcement learning for aerodynamic airfoil optimisation subject to structural constraints}

\author{David Ramos}

\affiliation{ETSIAE-UPM-School of Aeronautics, Universidad Politécnica de Madrid, Plaza Cardenal Cisneros 3, E-28040 Madrid, Spain}
\email{david.ramos.archilla@upm.es.}
\author{Lucas Lacasa}

\affiliation{Institute for Cross-Disciplinary Physics and Complex Systems (IFISC, CSIC-UIB), 07122 Palma de Mallorca, Spain}

\author{Eusebio Valero}

\affiliation{ETSIAE-UPM-School of Aeronautics, Universidad Politécnica de Madrid, Plaza Cardenal Cisneros 3, E-28040 Madrid, Spain}
\affiliation{Center for Computational Simulation, Universidad Politécnica de Madrid, Campus de Montegancedo, Boadilla del Monte, 28660 Madrid, Spain}

\author{Gonzalo Rubio}

\affiliation{ETSIAE-UPM-School of Aeronautics, Universidad Politécnica de Madrid, Plaza Cardenal Cisneros 3, E-28040 Madrid, Spain}
\affiliation{Center for Computational Simulation, Universidad Politécnica de Madrid, Campus de Montegancedo, Boadilla del Monte, 28660 Madrid, Spain}

\date{\today}

\begin{abstract}
The main objective of this paper is to introduce a transfer learning-enhanced deep reinforcement learning (DRL) methodology that is able to optimise the geometry of any airfoil based on concomitant aerodynamic and structural integrity criteria. 
To showcase the method, we aim to maximise the lift-to-drag ratio $C_L/C_D$ while preserving the structural integrity of the airfoil --as modelled by its maximum thickness-- and train the DRL agent using a list of different transfer learning (TL) strategies.
The performance of the DRL agent is compared with Particle Swarm Optimisation (PSO), a traditional gradient-free optimisation method. Results indicate that DRL agents are able to perform purely aerodynamic and hybrid aerodynamic/structural shape optimisation, that the DRL approach outperforms PSO in terms of computational efficiency and aerodynamic improvement, and that the TL-enhanced DRL agent achieves performance comparable to the DRL one, while further saving substantial computational resources.
\end{abstract}

\maketitle

%

\section{Introduction}
\vspace{-5mm}
Airfoil design is key to improving the performance and efficiency of various aerodynamic systems, such as aircraft wings, turbine blades, and wind turbines \cite{Lu2022}. Such design often predicates on (i) achieving an optimal balance between two aerodynamic forces --where one aims to maximise the aerodynamic lift and minimise the aerodynamic drag-- while (ii) maintaining structural integrity \cite{Bak_2014}. The aerodynamic balance is often explicitly modelled as the lift-to-drag ratio $C_L/C_D$, where $C_L$ and $C_D$ are the lift and drag coefficients. The optimisation problem can be formally stated as maximising an objective function $\mathscr{J}(x)$, with $x$ representing the airfoil design parameters.

\medskip \noindent 
The two common strategies for airfoil shape optimisation involve, respectively, using gradient-based or gradient-free methods. The former methods rely on the computation of $\nabla \mathscr{J}(x)$ to iteratively improve the design \cite{skinner2018state, garnier2019review}. 
This approach is computationally efficient --particularly with high-dimensional design spaces--, making it a popular choice in aerodynamics as highlighted by comparative studies \cite{secanell2005numerical, lyu2014benchmarking}. The well-known {\it adjoint method}, introduced in Fluid Dynamics by the seminal work of Pironneau \cite{pironneau1974optimum}, is one prominent example of a gradient-based technique; requiring a computational effort that scales independently of the number of design variables. Subsequently, this method was refined by the so-called  {\it discrete adjoint method}, becoming a widely used choice in this field \cite{frank1992comparison, jameson1998optimum}. While gradient-based methods are generally efficient and very popular in aerodynamic shape optimisation, they only have provable local convergence, and thus for non-convex objective functions their performance  
is sensitive to the initial design point \cite{jichao2022machine} and function continuity \cite{skinner2018state}.

\medskip \noindent 
On the other hand, gradient-free methods offer an alternative to complex, nonlinear optimisation problems that often appear in aerodynamic design \cite{skinner2018state, viquerat2021direct}. Despite their higher computational cost compared to gradient-based methods, they are more robust at finding global optima and handling challenging objective functions \cite{skinner2018state, garnier2019review}. Popular gradient-free methods deployed in the context of airfoil shape optimisation
include Genetic Algorithms (GA) \cite{bizzarrini2011genetic} or
Particle Swarm Optimisation (PSO) \cite{wang2011robust, fourie2002particle, nejat2014airfoil}. As advanced, these methods are often limited by their computational demands and can be inefficient with very large design spaces \cite{lyu2014benchmarking}. To partially bridge these issues, some recent efforts to improve the computational efficiency of gradient-free methods include, e.g., parallel implementations, which can make GAs viable even for intensive applications such as those in aerospace engineering \cite{wang2002parallel}.\\ Recently, deep reinforcement learning (DRL) \cite{arulkumaran2017deep} has emerged as a promising (gradient-free) tool for shape optimisation \cite{dussauge2023reinforcement, lou_aerodynamic_2023}. DRL methods have demonstrated remarkable capabilities in various domains, particularly when dealing with complex high-dimensional data \cite{garnier2019review}. Their impact in fluid dynamics studies and aeronautical applications belongs to a recently but rapidly expanding research field that encompasses both general-purpose ML-based fluid mechanics \cite{Brunton2020,Vinuesa2023,Vinuesa2022}, and their application to several aeronautical problems \cite{LeClainche2023, Brunton2021} with clear industrial impact \cite{lacasa2025towards}.

\medskip \noindent 
In the task of airfoil optimisation, a significant drawback of training DRL algorithms lies in the computational cost associated with solving the aerodynamic flow problem around the airfoil for a concrete shape: while the most accurate approach to solve the aerodynamic flow involves computational fluid dynamics (CFD) techniques, its intrinsic high computational cost requires opting for alternative, approximate solvers. For instance, some recent studies employ panel methods such as {\it XFoil} \cite{dussauge2023reinforcement, lou_aerodynamic_2023} to approximate the aerodynamic performance of airfoils, while some other studies \cite{liu_deep_2024, liu_airfoils_2023} focus on further reducing optimisation costs by constructing surrogate models instead. Although the latter offers faster training times and smoother reward variations with respect to the agent’s actions --facilitating the learning process--, these approximate models come at the price of a decrease in accuracy.

\medskip \noindent 
Interestingly, one significant advantage of recent deep learning methods in optimisation tasks is their potential to enhance their learning efficiency through so-called {\it Transfer Learning} \cite{YAN2019826, 10.1063/5.0134198}: the possibility of pre-training a given model on one task and subsequently fine-tuning it for a related but different task. Transfer learning could also be developed in the context of DRL, accelerating the agent's learning process by reusing elements such as policies or feature representations and effectively reducing the need for extensive exploration, a particularly valuable trait in aerodynamic optimisation. 
To the best of our knowledge this approach has been seldom explored, with a few encouraging exceptions for missile control surface and airfoils \cite{YAN2019826, BHOLA2023112018}. The novelty we bring with respect to these two works is a greater reduction of the cost when transfer learning is applied, and the thorough implementation of a suite of transfer learning-enhanced DRL framework for aerodynamic optimisation subject to structural constraints.\\ 
Additionally, note that most studies on airfoil shape optimisation focus solely on maximising aerodynamic efficiency \cite{liu_deep_2024, dussauge2023reinforcement, lou_aerodynamic_2023, YAN2019826, liu2025cnn}, without explicitly considering important structural integrity constraints \cite{buckley2010airfoil} related, e.g., to the airfoil’s area or thickness. Accordingly, there is a need to develop shape optimisation frameworks that aim to maximise the aerodynamic efficiency while preserving the structural integrity of the airfoil. This, we argue, can be interpreted as an example of multi-objective optimisation \cite{liu_deep_2024}, which in practice can be reduced to a single-objective one by suitable scalarisation.

\medskip To bridge these gaps, here we aim to explore the performance of DRL (enhanced with different possible Transfer Learning strategies) as a gradient-free framework for airfoil shape optimisation that considers the concomitant role of aerodynamic efficiency and structural integrity. 
To that aim, we define a flexible reward function that combines (i) the lift-to-drag ratio of the flow around the airfoil and (ii) the airfoil's maximum thickness. We
initially train our DRL agent using {\it XFoil} --both for aerodynamic optimisation alone, and combined aerodynamic shape optimisation with structural integrity preservation-- and show that the DRL agent is successful at both tasks. At this point results are compared with the ones obtained with Particle Swarm Optimisation, certifying the advantage of DRL over more traditional gradient-free methods. 
We then consider a Transfer Learning scenario where the DRL agent is pre-trained using {\it NeuralFoil} \cite{sharpe2025neuralfoil}, a faster-yet-less-accurate neural network surrogate model of {\it XFoil} built using domain knowledge. After an initial pre-training phase, we subsequently fine-tune the agent’s performance through transfer learning, incorporating direct training with {\it XFoil}. 
We compare four different transfer learning strategies and demonstrate how this combined approach can be successfully implemented, achieving accuracy levels comparable to those obtained through direct DRL training with {\it XFoil}, but at a reduced computational cost. 



\medskip \noindent 
The rest of the paper is organised as follows. In Section \ref{sec:methodology}, we introduce the whole methodology.  We start by formalizing the original multi-objective optimisation problem where aerodynamic and structural integrity criteria yield two different objective functions, and we suitably combine these into a reward function which effectively scalarises the problem into the framework of single-objective optimisation. Then, we introduce the reinforcement learning method and enhance it with several transfer learning strategies. In Section \ref{sec:results}, we report the results, comparing DRL agents with PSO (both) for pure aerodynamic shape optimisation and aerodynamic/structural integrity optimisation, trained with vs without transfer learning.
Finally, in Section \ref{sec:conclusions}, we conclude.

\section{Methodology}
\label{sec:methodology}
\subsection{Problem under study in a nutshell}
\vspace{-5mm}
We consider the two-dimensional geometric optimisation of an airfoil shape under fixed operating conditions. The conditions chosen for the study are an Angle of Attack $\text{AoA}=2$ degrees, a Mach number $\text{Ma}=0.5$, and a Reynolds number $\text{Re}=10^6$. These represent a typical aerodynamic scenario for subsonic flow, allowing us to investigate the optimisation of airfoil shapes under conditions that are common in practical applications. 
In general terms, we shall consider an airfoil shape to be `better' than another one when the aerodynamic flow around the airflow shows an enhanced lift-to-drag ratio. Accordingly, from an aerodynamic perspective the objective function to be maximised is such ratio, which is computed by applying an aerodynamic solver to any given airfoil geometry. Note, however, that aiming to optimise this ratio alone can yield geometries whose structural integrity can be compromised. Accordingly, in what follows we will also include structural constraints when we build the reward function.\\
In the next subsections, we detail our proposed geometric optimisation framework and its specifications. We will introduce a Reinforcement Learning paradigm which we enhance by incorporating a Transfer Learning strategy, where the agent suitably switches between using a fast, surrogate aerodynamic solver ({\it NeuralFoil} \cite{sharpe2025neuralfoil}) to a slightly slower, yet more accurate one ({\it XFoil} \cite{drela1989xfoil}) as the method to assess the aerodynamic flow and to compute the agent's reward. {\it XFoil} is
a classic model --based on a panel method coupled with a viscous boundary layer formulation-- designed for subsonic flows that calculates the pressure distribution on the airfoil and subsequently derives the lift and drag characteristics. It is faster than performing full DNS integration, but is still considered as an actual aerodynamic solver. {\it NeuralFoil} on the other hand --implemented in a Python-based library-- is a ML surrogate model that can provide representation of the viscous, compressible airfoil aerodynamics for nearly any airfoil, with control surface deflections, across a 360-degree angle of attack, at any Reynolds number. {\it NeuralFoil} has been trained with millions of {\it XFoil} simulations, enabling rapid calculations of aerodynamic characteristics, albeit with a slight trade-off in accuracy. Unlike {\it XFoil}, {\it NeuralFoil} avoids non-convergence issues. While training the agent with \textit{NeuralFoil} yields poorer results than when the agent is trained with \textit{Xfoil} (see Table \ref{tab:neuralfoil_fine-tuning_evaluation}), we will show that we can successfully leverage \textit{NeuralFoil} to reach optimal results in the context of Transfer Learning enhancement.\\
As a final note, the developed DRL geometric optimisation framework is available through the open source library pyLOM \cite{EIXIMENO2025109459}.  
\subsection{The basic DRL optimisation framework}
\begin{figure*}[htb!] 
    \centering 
    \includegraphics[width=0.8\textwidth]{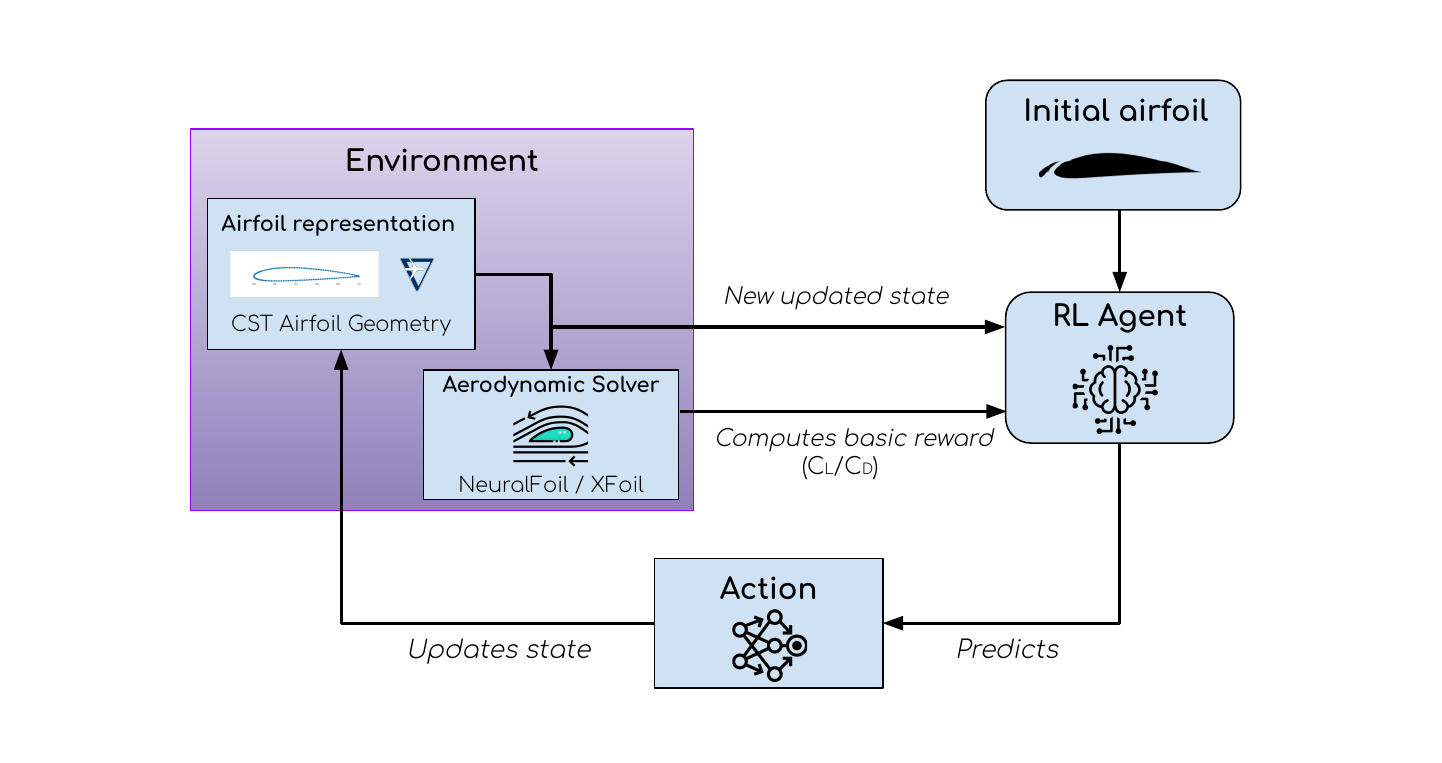}
    \caption{Basic DRL training diagram.} 
    \label{fig:RL_diagram}
\end{figure*}

\vspace{-5mm}
The initial aim is to develop a DRL framework, where the DRL {\it agent} is capable of improving the aerodynamic efficiency of a given airfoil, as illustrated in Fig.~\ref{fig:RL_diagram}. To that aim, the agent interacts with the {\it environment} (the airfoil geometry, which produces a certain aerodynamic flow) and its goal is to find a {\it policy} that ultimately maximises the lift-to-drag ratio ($C_L/C_D$) of such aerodynamic flow, by adequately performing {\it actions} that adjust the design variables. 




\subsubsection{RL specs I: environment, state space and action definition}
\vspace{-5mm}
\noindent {\bf Environment.} The initial stage of the process involves the creation of an {\it environment} in which the agent can undergo training --see Fig.~\ref{fig:RL_diagram}--, and includes (i) choosing an adequate airfoil representation, and (ii) choosing an aerodynamic solver that will compute the aerodynamic properties of the airfoil. Regarding (i), observe that a priori one could simply use the 2D
coordinates of the points over the airfoil. However, this choice is delicate as it may lead the agent to explore unusual or even unphysical shapes. A better choice is to use the so-called Class-Shape Transformation (CST) parameters \cite{kulfan2008universal}, a widely-used technique to parametrise geometries based on the curvature of the shape along the chord. This parameterisation indeed generates more realistic shapes and reduces the dimensionality of the design space, allowing for a more efficient optimisation.\\
The implementation of such environment was done using the Gymnasium Python library \cite{towers2024gymnasium}, a maintained fork of the OpenAI Gym library, which provides a flexible framework for defining and managing RL environments. 

\medskip
\noindent {\bf State space.} 
The state space spans a total of 18 scalar parameters $\textsc{State}=(s_1,s_2,\dots,s_{18})$, where each parameter is bounded in a concrete range $s_i \in [\textsc{LowerBound}(i), \textsc{UpperBound}(i)]$. \textsc{LowerBound} and \textsc{UpperBound} are two vectors that define the boundaries of the state space and are defined in Appendix~\ref{ap:env_configuration}.
These 18 parameters are converted to an airfoil using CST representation, such that eight of them characterise the upper surface, eight of them the lower surface, one for the leading-edge weight, and one for the trailing-edge thickness. 
Different configuration values of these 18 parameters within their ranges correspond to different airfoil geometries.

\medskip
\noindent {\bf Action space and state update.} To modify the state of the environment (the airfoil), the agent predicts an action, i.e. a vector $\textsc{Action}=(\textsc{action}(1), \textsc{action}(2), \dots, \textsc{action}(18))$ where $\textsc{action}(i)\in[-1,1] \ \forall i$, as recommended in the documentation of stable baselines\footnote{https://stable-baselines3.readthedocs.io/en/master/guide/rl\_tips.html\#tips-and-tricks-when-creating-a-custom-environment}.
This vector is the one performing the change in the parameters that represent the airfoil. Now, to prevent the sequential changes to the airfoil from being too abrupt, we further define a weight vector $\alpha=(\alpha_1,\alpha_2,\dots,\alpha_{18})$, where
\[
\alpha_i = \frac{\textsc{UpperBound}(i) - \textsc{LowerBound}(i)}{\textsc{EpisodeMaxLength}},  
\]
where \textsc{EpisodeMaxLength} is a positive integer that represents the maximum permitted length of an episode, defined with more detail in Sec. \ref{sec:episode}. This allows the agent to potentially explore both limits of the state space in a single episode, regardless of the initial state.
Accordingly, updating the state based on the action follows
\begin{equation}
\textsc{state}(i) \xleftarrow{} \textsc{state}(i) + \alpha(i) \cdot \textsc{action}(i), \ i=1,2,\dots,18
\label{eq:newstate_scalar}
\end{equation}
or in vector form
\begin{equation}
\textsc{State} \xleftarrow{} \textsc{State} + \alpha \odot \textsc{Action},
\label{eq:newstate}
\end{equation}
where $\odot$ is the element-wise multiplication (Hadamard product). 

\subsubsection{RL specs II: Episode definition}
\label{sec:episode}
\vspace{-5mm}
A single step of the DRL agent consists in, given an action, updating the airfoil geometry according to Eq.~\ref{eq:newstate}, assessing how this new geometry improves (or not) the aerodynamic efficiency and checking if the termination condition is met. We have decided that this condition will be reached when a maximum number of steps has been completed or if the aerodynamic solver, in this case \textit{XFoil}, does not converge.
The iteration of many steps until the termination condition defines an {\it episode}, and the number of steps taken is called the episode length.\\ 
Here, we initially define the maximum length of an episode as $\textsc{EpisodeMaxLength}=100$. Observe that this length needs to strike a balance: if it is too large, 
the agents will have problems to converge since this would reduce the number of episodes during training and the agents would receive less varied data. Moreover, the number of possible states increases, making the problem more complex.
On the other hand, if the length is too short, the values of $\alpha$ in Eq.~\eqref{eq:newstate} will be comparatively too large and the airfoil modifications could be too abrupt, further causing instability in the optimisation process.





\medskip
\noindent 
Once an episode is defined, we need to define how the start of a new episode is carried out. A standard approach is to fix the initial state (to a certain airfoil geometry, e.g. \texttt{naca0012}), such that at the beginning of each episode the state of the environment is always reset to the same airfoil. This is, for example, the choice in \cite{dussauge2023reinforcement}. Some other studies \cite{lou_aerodynamic_2023} in turn decide to reset to a different airfoil geometry, the rationale being that in this way the agent is capable of learning to optimise starting from a variety of shapes, i.e. the learned policy is more universally applicable. Here we chose the latter strategy, so that at the beginning of an episode the environment is chosen at random from a pre-compiled set of 20 \texttt{naca} airfoils
listed in the Appendix \ref{ap:env_configuration}. 

\subsubsection{RL Specs III: Reward function} 
\vspace{-5mm}
Once the episode has been defined, we need to specify the reward function, which estimates the quality of any given action. For pedagogical reasons, we now outline the construction of such reward function in a step-by-step way. As a starting point, we consider the maximisation of the lift-to-drag ratio $C_L/C_D$ --where both aerodynamic coefficients are computed from the aerodynamic solver--, which yields a purely aerodynamic shape optimisation.\\
As previously mentioned, in this work we will use two solvers: {\it XFoil} and {\it NeuralFoil}. Observe that, in addition to calculating lift and drag coefficients, {\it NeuralFoil} also returns a `confidence' value: a scalar $\kappa \in [0,1]$ that indicates how confident the surrogate model is after making the lift and drag predictions. In order to penalise weird geometries or geometries whose computed aerodynamic properties have large uncertainty, we will aim to maximise $\kappa\cdot (C_L/C_D)$, i.e. we aim at finding geometries that at the same time are (i) improving the lift-to-drag ratio and (ii) keeping the confidence high. With this term $\kappa$, we effectively mitigate the risk of the agent relying on potentially inaccurate predictions in regions where {\it NeuralFoil} is known to perform poorly.\\ 
Now, since in RL the agent tries to maximise the cumulative reward \cite{dussauge2023reinforcement}, the actual reward function needs to take a differential form, e.g. 
\begin{equation}
   \kappa_i\frac{C_L}{C_D}\bigg\vert_i - \kappa_{i-1}\frac{C_L}{C_D}\bigg\vert_{i-1}. \label{eq:pure}
\end{equation}

\begin{figure}[h]
    \centering
    \includegraphics[width=0.8\linewidth]{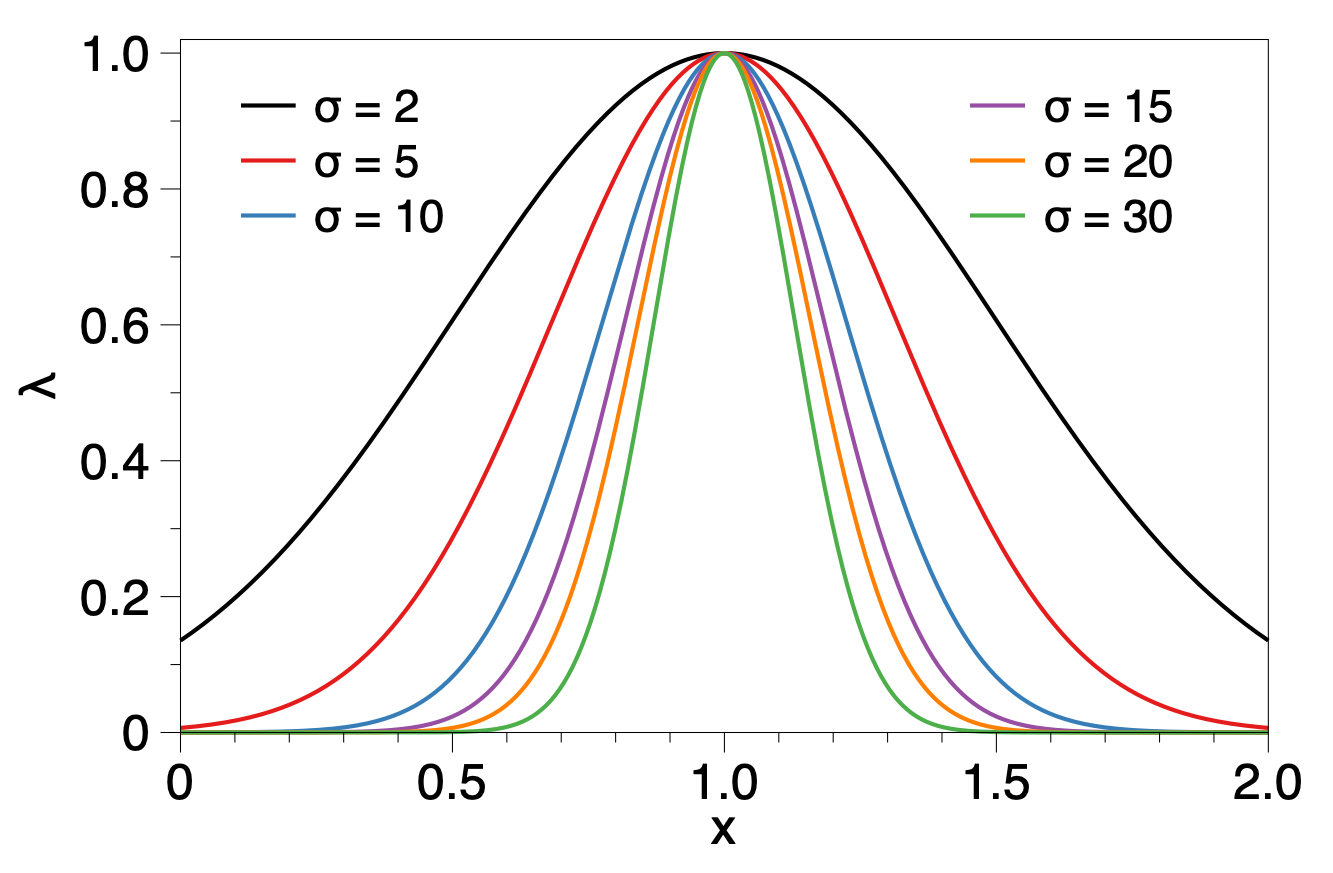}
    \caption{The Gaussian regularisation term $\lambda = \exp(-\sigma[x-1]^2)$ as a function of the hyperparameter $\sigma$. This term penalises geometries where the airfoil's maximum thickness deviates from the initial one, as captured by a maximum thickness quotient $x$ that deviates from one.}
    \label{fig:comparacion_sigmas}
\end{figure}

\medskip
\noindent So far, as previously mentioned the reward function denotes a purely aerodynamic optimisation. So at this point we consider as a second objective function the needs to preserve the structural integrity of the airfoil. As a simply proxy, we consider the airfoil's
maximum thickness $\textsc{mt}$, and our goal is to preserve as much as possible this structural property, i.e. we aim to minimise the mismatch between the initial and final thicknesses. We formalise this by using a Gaussian kernel $\lambda$, such that at step $i$, the kernel reads 
\begin{equation}
    \lambda_i = e^{-\sigma(x_i-1)^2},
    \label{eq:area_factor}
\end{equation}
where 
$x_i:= \textsc{mt}_i/ \textsc{mt}_0$ is the ratio between the airfoil's maximum thickness at step $i$ and the initial one, and $\sigma >0$ is an hyperparameter that controls the strength of the nonlinear penalisation. 
This Gaussian regularisation is inspired by computational fluid mechanics mesh optimisation methods that minimise error while maintaining low computational cost \cite{huergo2024reinforcementA,huergo2024reinforcementB}.
For a fixed $\sigma$, $\lambda$ simply penalizes those airfoils whose maximum thickness is not preserved (see Fig.~\ref{fig:comparacion_sigmas} for an illustration of how the values of $\lambda$ penalises geometries with non-conserved maximum thickness, for different values of $\sigma$). As such we now have to maximize the term given in Eq.~\ref{eq:pure} and, independently,  maximize $\lambda$ in Eq.~\ref{eq:area_factor}. This, in essence, is a multi-objective optimisation problem: we aim to find airfoil geometries that maximise the aerodynamic properties and maximally preserve the initial maximum thickness (a proxy for preserving the structural integrity). Now, instead of treating the aerodynamic and the structural aspects independently, we choose to scalarize the multi-objective problem by combining the aerodynamic and the structural terms into a single reward function. 
Finally, we can define such reward at step $i$ as
\begin{equation}
    \mathscr{R}_i = \lambda_i\kappa_i\frac{C_L}{C_D}\bigg\vert_i - \lambda_{i-1}\kappa_{i-1}\frac{C_L}{C_D}\bigg\vert_{i-1}, \ \mathscr{R}_0 := \kappa_{0}\frac{C_L}{C_D}\bigg\vert_{0}.
    \label{eq:reward_with_condifence_definitive}
\end{equation}
Note that when we use {\it XFoil} then $\kappa$ is not defined, in that case we set $\kappa=1$ in Eq.~\ref{eq:reward_with_condifence_definitive}. Since we now have a single reward function, observe that from a mathematical point of view we have approximated a multi-objective optimisation problem with a simpler, single-objective one which balances aerodynamic and structural integrity constraints. The Gaussian kernel effectively acts as a structural regularisation term.\\
In what follows, we will distinguish the special case $\sigma=0$ (yielding $\lambda_i=1, \forall i)$ which will be called the {\it purely aerodynamic} optimisation, from the more general case $\sigma\ne0$ where structural integrity enters in the reward function. This latter scenario will be called the {\it hybrid aerodynamic/structural} optimisation case, although we need to make clear at this point that optimisation in the structural dimension is simply aimed at minimising the change in maximum thickness.

\subsubsection{RL specs IV: the agent}
\begin{figure}[htb!]
    \centering
    \includegraphics[width=1\linewidth]{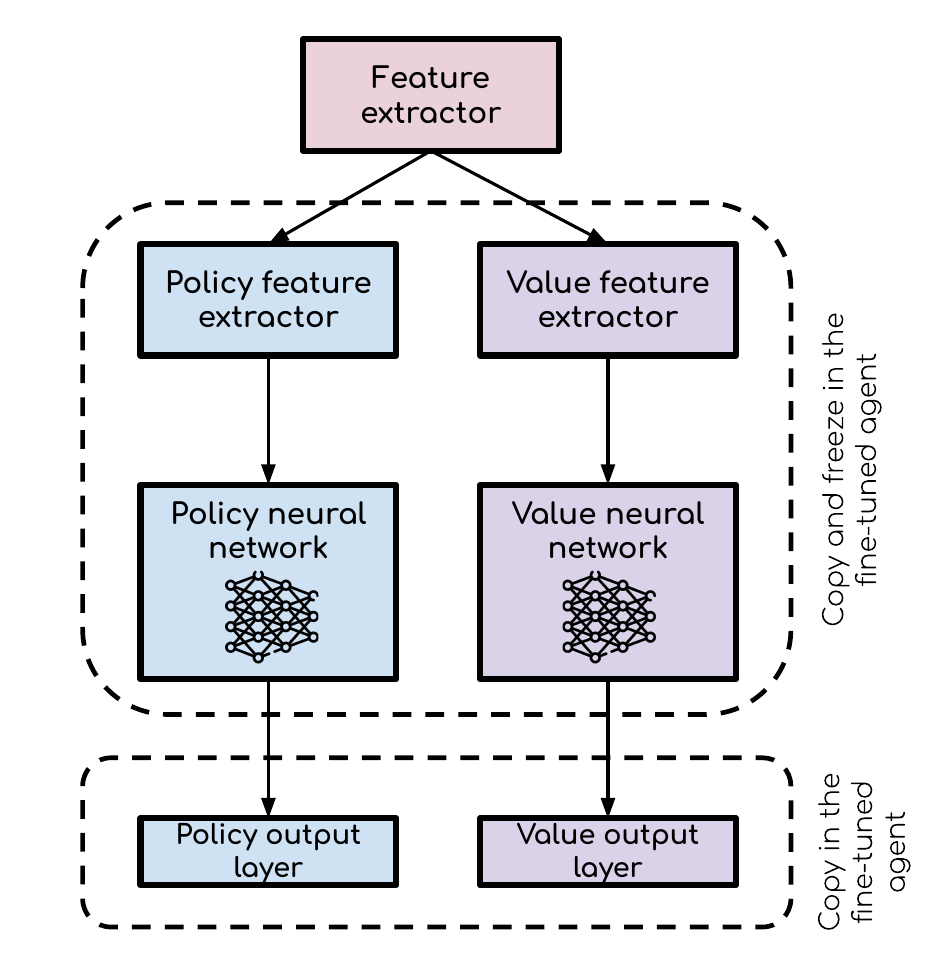}
    \caption{Architecture of the DRL agent that uses Proximal Policy Optimisation. The figure illustrates the transfer learning strategy $\# 3$, highlighting the part of the agent's parameters which are copied and frozen in the fine-tuning phase and the part which is copied and is still trained in during fine-tuning.}
    \label{fig:transfer_learning_ppo}
\end{figure}

\vspace{-5mm}
To train an agent with reinforcement learning, one needs to choose the specific DRL algorithm. This choice may depend, among other things, on whether the action space is discrete or continuous.
For continuous actions, there are a variety of algorithms, including 
A2C \cite{mnih2016asynchronous}, PPO \cite{schulman2017proximalpolicyoptimizationalgorithms}, SAC \cite{haarnoja2018soft} or TD3 \cite{fujimoto2018addressing}. 
 After running some preliminary analysis, we decided to use PPO (Proximal Policy Optimisation), as it shows better convergence properties than A2C. Moreover, we discarded the use of TD3 and SAC since they had difficulties converging. PPO is a popular policy gradient algorithm that, for instance, is behind the post-training with RLHF of popular Large Language Models, underscoring its versatility.
In our implementation, PPO employs neural networks to model the policy and value functions. Specifically, PPO uses an actor-critic framework where the policy network (actor) selects the actions, and the value network (critic) evaluates the expected returns of those actions. Figure \ref{fig:transfer_learning_ppo} shows a schematic of the neural networks comprising PPO.
To train the agent, the \texttt{Python} library \texttt{Stable Baselines 3} \cite{stable-baselines3} has been used. This is a library that implements different reinforcement learning algorithms and is easy to integrate into \texttt{Gymnasium} environments. 

\subsection{Transfer learning}
\label{sec:transfer_learning}
\vspace{-5mm}
Finally, we need to explicitly discuss the transfer learning technique. We start by recalling that this is a way of accelerating the training process \cite{YAN2019826}, as it has been observed in RL frameworks that an agent initially trained with a (fast) low-cost model over a substantial number of iterations and subsequently trained with a high-cost model over a limited number of iterations achieves a comparable performance with respect to training solely
with a high-cost model over a longer period. Transfer learning thus reduces the overall computational cost and time required for the training process, something particularly welcome in high-cost applications such as aerodynamic shape optimisation. Here, as previously discussed, we use the surrogate model {\it NeuralFoil} as the low-cost model, and {\it XFoil} as the more accurate yet more computationally expensive solver. To provide a more effective evaluation of the efficacy of Transfer Learning we use the smallest model available of {\it NeuralFoil}, which has a clear speed advantage and a clearly lower accuracy compared to {\it XFoil}.

\medskip \noindent In practice, we fine-tune the actor-critic neural network utilised by the PPO algorithm. Fine-tuning is a common technique in machine learning where a pre-trained model is adapted to a new task or domain by continuing the training process with different data. Accordingly, here the initial step is to train an agent with the low-cost model. Subsequently, the weights of this agent are transferred to a new agent that is then trained with the high-cost model. Several different approaches have been considered: 
\begin{itemize}
    \item $\# 1:$ Share all the parameters of the network and continue with the training.
    \item $\# 2:$ Share all parameters except those of the last layer, both for the action network and the value network, and then continue with the training.
    \item $\# 3:$ Share all the parameters, freeze all of them except those of the last layer of the network, and continue the training.
    \item $\# 4:$ Share all the parameters except those of the last layer of the network, both for the action network and the value network, freeze them, and continue the training.
\end{itemize}
To clarify, by `freezing of weights' during fine-tuning we mean that they will not be changed. Fig.~\ref{fig:transfer_learning_ppo} illustrates how the weight transfer is performed for the specific scenario raised on strategy $\#3$. Finally, note that after pre-training the first agent, an entropy term is added to the PPO loss function, which encourages the new model to explore the state space. This makes it easier for this new agent to move into other areas of the state space. The PPO configurations for each case are described in Appendix \ref{ap:ppo_params}.

\section{Results}
\label{sec:results}
\vspace{-3mm}
Results cover a variety of analysis of the DRL agents, see Appendix \ref{ap:ppo_params} for an in-depth study of the hyperparameter selection and the final configurations of the different PPO hyperparameters.\\ 
We initially compare the performance and training's computational efficiency of the DRL agent (trained for the purely aerodynamic optimisation and then also trained on hybrid aerodynamic/structural optimisation) with respect to a traditional gradient-free optimisation methodology, namely Particle Swarm Optimisation (PSO). Then we perform a more detailed analysis of the performance of the DRL agent in the single vs multi-objective scenario. Finally, we assess the performance and efficiency gain when the DRL agent is enhanced with different possible Transfer Learning strategies.


\medskip \noindent
We recall that each episode has at most 100 steps and that training airfoils are sampled randomly at the beginning of each episode from the set of 20 airfoils specified in previous sections. On top of the computing the reward in every step following Eq.~\ref{eq:reward_with_condifence_definitive}, we monitor the average episode's reward each \textsc{n\_steps}, see Appendix \ref{ap:ppo_params} for details. The episode reward is defined simply as the sum of the rewards $\mathscr{R}_i$ of each step $i$ in an episode. The criterion to finish training is based on observing a plateau in this metric.\\
Once the DRL agent is trained, we evaluate its performance by running a whole episode on the set of airfoils in the Aerosandbox \cite{aerosandbox_masters_thesis} Python library, which is essentially a superset of the UIUC Airfoil dataset. This dataset includes a broad variety of airfoils, which allows for a comprehensive evaluation of the agent’s ability to generalise across different shapes and aerodynamic characteristics. 
In the evaluation, we always use {\it XFoil} as the method to compute the lift-to-drag ratios, so as to have a fair way of comparing agents. Now, since {\it XFoil} is known to sometimes have convergence issues, the total number of airfoils evaluated fluctuates a bit between agents.\\
The evaluation metrics include the best lift-to-drag ratio attained within each episode $$\texttt{best}=\max \bigg \{\frac{C_L}{C_D}\bigg\vert_i\bigg \}_{\text{episode}},$$
 and the net improvement (in lift-to-drag ratio units) within each episode
$$\texttt{improvement}= \texttt{best} - \frac{C_L}{C_D}\bigg |_{\text{initial}}.$$
Aditionally, in the hybrid optimisation case we monitor the deviation of the airfoil's maximum thickness with the percentage
$$\Delta \textsc{mt}=\frac{|\textsc{mt}_{\text{opt}}-\textsc{mt}_{\text{ini}}|}{\textsc{mt}_{\text{ini}}}\cdot 100,$$
where  $\textsc{mt}_{\text{ini}}$ and $\textsc{mt}_{\text{opt}}$ are the maximum thickness of the initial airfoil and the airfoil with best lift-to-drag ratio, respectively.

\subsection{Deep Reinforcement Learning {\it vs} Particle Swarm Optimisation}
\vspace{-5mm}
This section assesses the proposed DRL methodology and compares it (both for purely aerodynamic and hybrid optimisation) with a traditional gradient-free method: Particle Swarm Optimisation (PSO) \cite{kennedy1995particle}. 
PSO is a population-based optimisation algorithm inspired by the social behaviour of bird flocks searching for corn. It is widely used due to its simplicity, ability to handle non-differentiable objective functions, and efficiency in exploring high-dimensional search spaces. PSO has been successfully applied in various domains, including image and video analysis applications, engineering design and neural networks, and fluid dynamics optimisation. Furthermore, PSO has proven effective in airfoil optimisation \cite{wang2011robust, fourie2002particle, nejat2014airfoil} and seems to better suited than other standard gradient-free choices such as Genetic Algorithms for airfoil optimisation \cite{mukesh2012influence, hoyos2021airfoil}.\\
Here, we use {\it Xoptfoil2} \cite{Xoptfoil2} to perform PSO optimisation, since this tool is well designed to compare PSO and DRL as it actually makes calls to {\it XFoil} underneath.
The chosen PSO parameters are the default ones that this library provides. The results obtained using this tool are systematically compared to those given by the DRL agent trained for 81920 steps using {\it XFoil}. Since the main point of this paper is DRL, we do not conduct a thorogh hyperparameter optimisation for PSO, so we cannot guarantee that its results may slightly differ after thorough hypertuning.
The configuration of {\it XFoil} used in the training of the DRL agent and the PSO optimiser is detailed in Table~\ref{tab:xfoil_config}.


\begin{figure}[htb!]
    \centering
    \begin{subfigure}{\linewidth}
        \centering
        \includegraphics[width=1\linewidth]{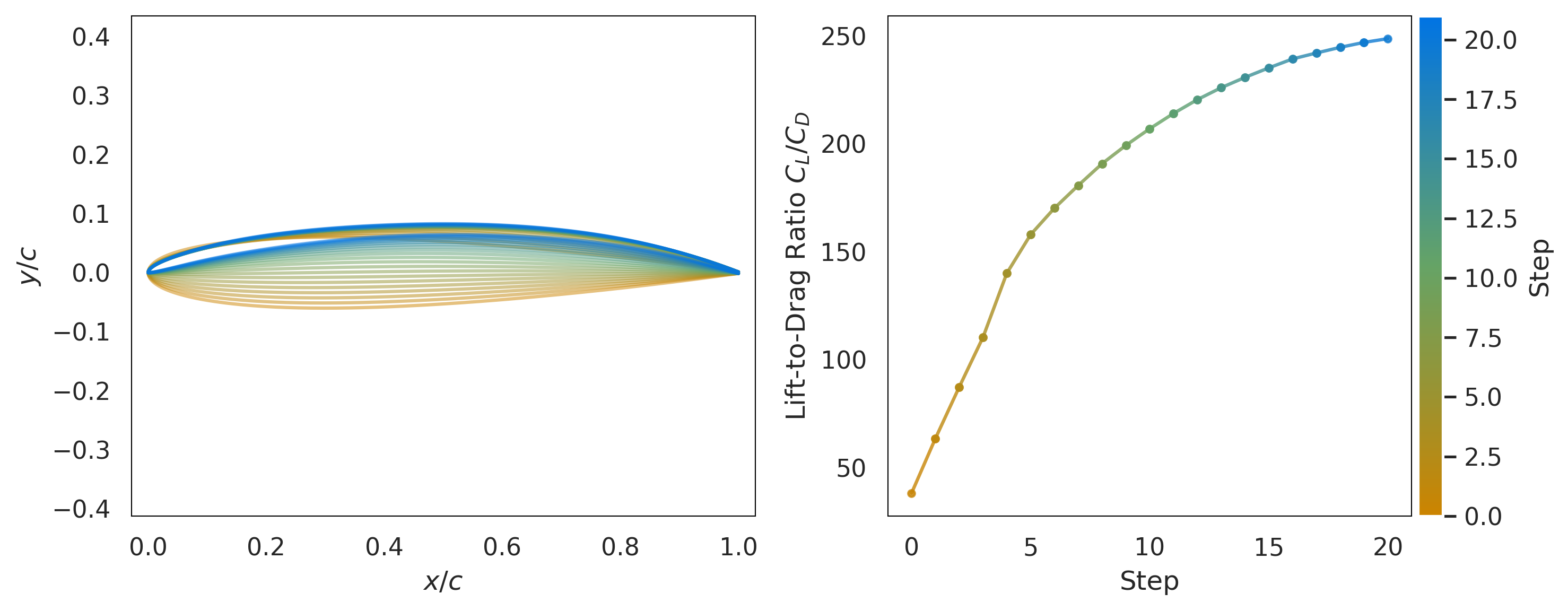}
        \caption{Optimisation of a \texttt{NACA0012} airfoil by a DRL agent trained on purely aerodynamic optimisation. Here, x/c and y/c mean the x and y coordinates of the airfoil normalized to the chord length.}
        \label{fig:subfig1}
    \end{subfigure}
    \begin{subfigure}{\linewidth}
        \centering
        \includegraphics[width=1\linewidth]{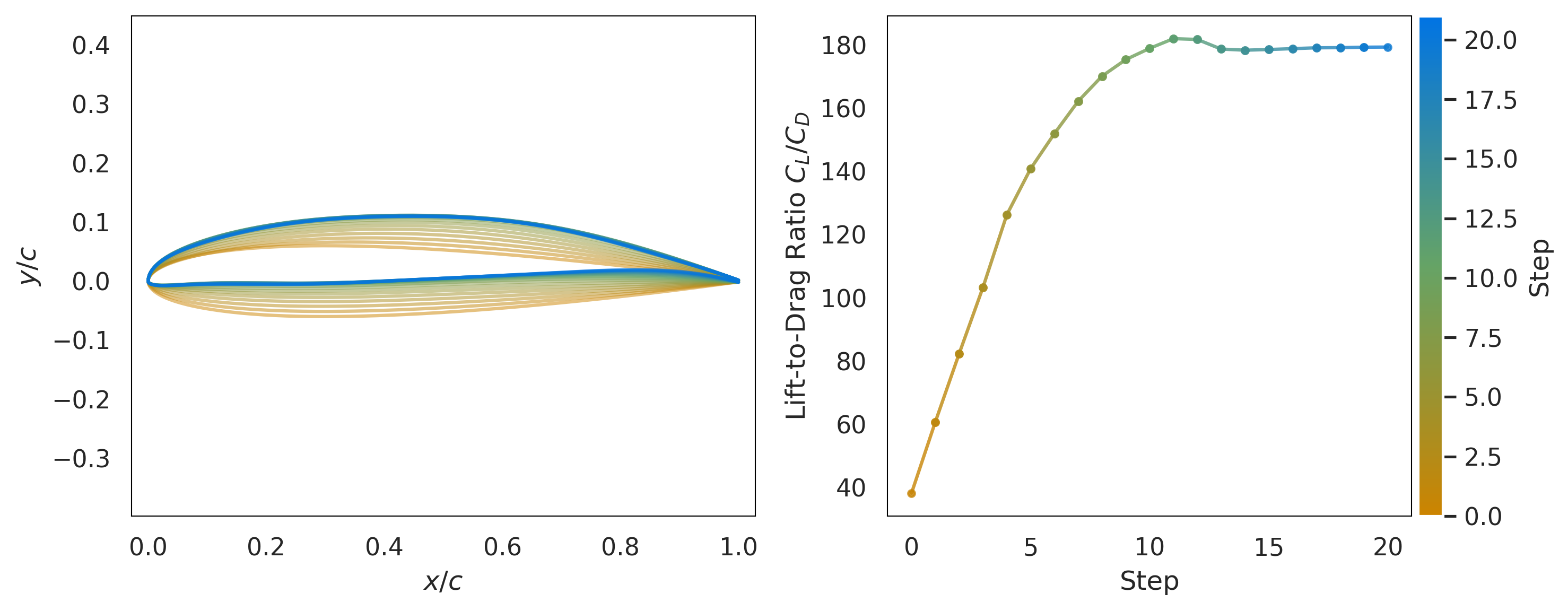}
        \caption{Optimisation of a \texttt{NACA0012} airfoil by a DRL agent trained on hybrid aerodynamic/structural optimisation ($\sigma=15$). Here, x/c and y/c mean the x and y coordinates of the airfoil normalized to the chord length.}
        \label{fig:subfig2}
    \end{subfigure}
    \begin{subfigure}{\linewidth}
        \centering
        \includegraphics[width=1\linewidth]{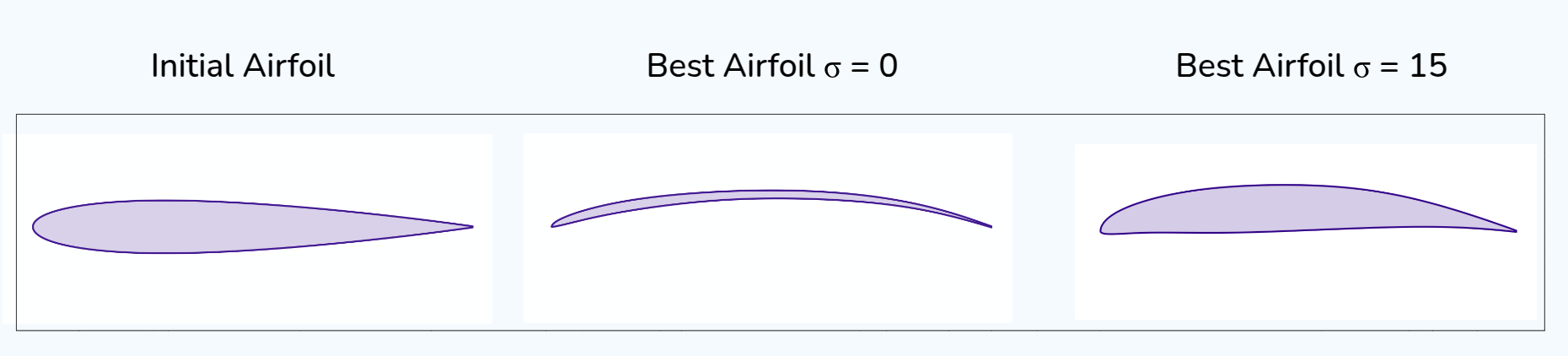}
        \caption{Left: initial \texttt{NACA0012} airfoil. Center: Purely aerodynamic optimisation. Right: Hybrid aerodynamic/structural optimisation ($\sigma=15$).}
        \label{fig:subfig3}
    \end{subfigure}
    \caption{Illustration of a DRL policy (20 steps) on \texttt{NACA0012} trained on aerodynamic and aerodynamic/structural optimisation, respectively.}
    \label{fig:comparison_penalization_area}
\end{figure}

\medskip
We start with the purely optimisation case, where the DRL agent is trained to solely improve the aerodynamic efficiency of the airfoil. For illustration, the first 20 steps of the DRL policy, as applied to the initial airfoil \texttt{NACA0012}, is shown in Fig.~\ref{fig:comparison_penalization_area}a, whereas the airfoil reaching the highest lift-to-drag ratio is depicted in Fig.~\ref{fig:comparison_penalization_area}c (center). We clearly see that the DRL agent's policy is able to substantially and monotonically increase the aerodynamic efficiency of the airfoil.\\
Now, to initialy compare the performance of DRL and PSO optimisation, we select 20 airfoils at random from the UIUC dataset --checking that they represent a wide variety of shapes-- and subsequently applied the policies from DRL and PSO. 
The initial and final $C_L/C_D$ results of these trajectories are shown in Table~\ref{tab:airfoil_comparison_pso_rl}. We find that DRL outperforms PSO in most of the cases, certifying the suitability of DRL for airfoil shape optimisation and suggesting its supremacy over PSO.\\
Then, we tackle the whole UIUC airfoil dataset and proceed to make the full comparison between DRL and PSO. 
The performance metrics \texttt{best} and \texttt{improvement} (averaged over the whole set of evaluated airfoils) are depicted in Table~\ref{tab:optimisation_comparison_uiuc}, certifying the supremacy of DRL over PSO for aerodynamic optimisation.
\begin{table}[htbp]
   \centering
   \begin{tabular}{|l|c|c|c|c|c|c|}
       \hline
       \textbf{Airfoil} & \textbf{Initial} & \textbf{PSO single} & \textbf{DRL single} & \textbf{PSO multi} & \textbf{DRL multi}\\
       \hline
       \texttt{naca0012} & 38.32 & 47.72 & \textbf{249.4} & 45.91 & \textbf{103.6} \\
       \texttt{naca2412} & 80.53 & 216.7 & \textbf{257.1} &162.2 & \textbf{184.1}\\
       \texttt{s1223} & 93.69 & \textbf{239.1} & 204.4 & \textbf{156.2} & 115.0\\
       \texttt{fx63143} & 76.12 & 86.67 & \textbf{217.4} & 63.57 & \textbf{102.1}\\
       \texttt{e387} & 119.9 & 209.2 & \textbf{224.8} &194.9 & \textbf{197.3}\\
       \texttt{naca4412} & 113.0 & 223.0 & \textbf{250.7} &\textbf{163.4} & 141.0 \\
       \texttt{\shortstack{naca63\\-2a015}} & 54.34 & \textbf{158.8} & 76.5 & \textbf{117.9} & 104.8\\
       \texttt{clarky} & 107.2 & 215.1 & \textbf{225.5} & 164.5 & \textbf{181.7}\\
       \texttt{e174} & 122.0 & 218.7 & \textbf{231.3} & 194.4 & \textbf{197.0}\\
       \texttt{mh113} & 116.9 & 247.1 & \textbf{254.9} &145.3 & \textbf{169.4}\\
       \texttt{ag25} & 102.7 & \textbf{250.5} & 249.8 &\textbf{215.1} & 215.0\\
       \texttt{raf48} & 71.33 & 201.5 & \textbf{248.4} &138.0 & \textbf{159.1}\\
       \texttt{naca23015} & 43.53 & 146.0 & \textbf{257.3} &86.30 & \textbf{167.7}\\
       \texttt{sc20012} & 31.69 & 51.24 & \textbf{201.0} &45.19 & \textbf{157.0}\\
       \texttt{goe398} & 100.5 & 240.5 & \textbf{257.1} &142.6 & \textbf{170.7}\\
       \texttt{n64212} & 75.90 & 222.5 & \textbf{261.8} &\textbf{166.4} & 133.0\\
       \texttt{ls013} & 44.53 & 52.51 & \textbf{231.6} &45.61 & \textbf{168.0}\\
       \texttt{dae21} & 155.1 & 203.3 & \textbf{258.0} &\textbf{259.8} & 177.8\\
       \texttt{vr7} & 94.59 & 236.3 & \textbf{253.6} &153.0 & \textbf{187.0}\\
       \texttt{\shortstack{naca66\\-2415}} & 93.72 & 199.7 & \textbf{240.8} &147.5 & \textbf{160.7}\\
       \hline
   \end{tabular}
   \caption{Comparison of PSO and DRL performance for purely aerodynamic and hybrid aerodynamic/structural optimisation (in terms of the highest $C_L/C_D$ reached in each of the two optimisation paradigms), across a set of different airfoil with different initial $C_L/C_D$. On average, DRL reaches better aerodynamic efficiency than PSO.}
   \label{tab:airfoil_comparison_pso_rl}
\end{table}


\begin{table}[h]
\centering
\begin{tabular}{|l|c|c|c|}
\hline
\textbf{Method} & \textbf{\shortstack{Evaluated \\ Airfoils}} & \textbf{Improvement} & \textbf{Best} \\
\hline
PSO (pure aerodynamic) & 1566 & 105 $\pm$ 50 & 
198(76) \\
\hline
DRL (pure aerodynamic) & 1456 & 141 $\pm$ 48 & 
243(35)\\
\hline
\end{tabular}
\caption{Comparison of PSO and DRL evaluation for purely aerodynamic optimisation  with the whole UIUC airfoil dataset, cerifying the supremacy of DRL over PSO for purely aerodynamic optimisation. The \texttt{Improvement} metric is provided in terms of mean $\pm$ one standard deviation (when averaged over the whole UIUC ensemble), whereas the \texttt{best} metric is provided in terms of median (IQR), since its distribution is left-skewed.}
\label{tab:optimisation_comparison_uiuc}
\end{table}

In a second step, we then focus on the hybrid optimisation case. For illustration, in Fig.~\ref{fig:comparison_penalization_area}b we show the performance of a DRL agent's policy which was trained on hybrid aerodynamic/structural optimisation with $\sigma=15$ (see also Fig.~\ref{fig:comparison_penalization_area}c (right)). The choice of $\sigma=15$ is arbitrary, it is not aimed at signalling any Pareto optimality but rather is selected for illustration purposes. The interested practitioner will actually need to select the value of $\sigma$ depending on the specific priorities of their application.\\ It is clear that the incorporation of the structural regularisation in the reward function has an impact: the lift-to-drag ratio is still increased (albeit reaching smaller values than in the purely aerodynamic case), but now the maximum thickness of the shape seems to be maintained.\\
Interestingly, within {\it Xoptfoil2} it is possible to define geometrical constraints --such as e.g. a target maximum thickness-- and therefore a fair comparison between DRL and PSO within the hybrid optimisation case discussed in Sec.~\ref{sec:multi} is possible too. For illustration, in this comparison we set $\sigma=15$ (that we will later show to  achieve a good balance between aerodynamic optimisation and maximum thickness conservation). 
A summary of the performance of the 20 previously selected evaluation airfoils can also be seen in Table \ref{tab:airfoil_comparison_pso_rl}, 
whereas summary statistics of the evaluation over all the converged airfoils of the UIUC airfoil dataset are reported in Table~\ref{tab:optimisation_comparison_uiuc_with_thickness}. It is evident that DRL exhibits superior performance in terms of aerodynamic efficiency optimisation. However, PSO demonstrates a more effective capability in preserving the maximum thickness. It's possible that this might just be flagging the capacity of DRL to explore more design space than PSO, that in that sense only optimises locally, i.e. performs more exploitation.\\
One thing to note from Table \ref{tab:airfoil_comparison_pso_rl} is that PSO anecdotically optimises certain airfoils more effectively in both pure and hybrid optimisation scenarios. One possible reason for the behavior in these cases is that if during optimisation with DRL one of the early states of the episode did not converge, making the modifications that the agent does to the optimised airfoil too small. This is, for instance, what happens with NACA63-2a015. Another reason why DRL may perform worse than PSO is because it looks for unrealistic shapes, erring on the side of creativity. This is what appears to happen with airfoil S1223. Finally, there may be airfoils that PSO simply optimises better.

\begin{table}[h]
\centering
\begin{tabular}{|l|c|c|c|c|}
\hline
\textbf{Method} & \textbf{\shortstack{Evaluated \\ Airfoils}} & \textbf{Improvement} & \textbf{Best} & {$\Delta$\textsc{mt} (\%)}\\
\hline
PSO (hybrid) & 1566 & 64 $\pm$ 36 & 
150(61) & 0.67 $\pm$ 2.21 \\
\hline
DRL (hybrid) & 1444 & 89 $\pm$ 37 & 
179(46) & 11.5 $\pm$ 9.4 \\
\hline
\end{tabular}
\caption{Comparison of 
of PSO and DRL performance for hybrid aerodynamic/structural optimisation ($\sigma=15$)
on the whole UIUC airfoil evaluation dataset. Whereas DRL clearly allows to reach higher aerodynamic efficiency, PSO makes a better job at preserving the maximum thickness.
Pointwise statistics of the \texttt{Improvement} and \texttt{best} metrics are provided as in Table \ref{tab:optimisation_comparison_uiuc}.}
\label{tab:optimisation_comparison_uiuc_with_thickness}
\end{table}

Finally, we provide a comparison of optimisation time per airfoil (averaged over the whole UIUC evaluation set) for DRL and PSO, as performed
on an Intel(R) Xeon(R) Gold 6248R CPU @ 3.00GHz+.
For concreteness, we focus on the purely aerodynamic case
(where the maximum thickness is not necessarily preserved, e.g. $\sigma=0$), although the results are qualitatively similar in the hybrid optimisation case. The DRL agent, using 1 process and 1 CPU thread, takes about 23 seconds to optimise all 1566 airfoils, i.e. an average of 0.0147 s/airfoil.
Comparatively, PSO was notably slower; actually, too slow to directly perform the optimisation in a single core, so MPI was used to accelerate the evaluation (and then the results per process were inferred). The CPU used has 48 cores, 47 used by worker processes plus the one by master process, and with this setup, optimising all the evaluation set took 22 hours and 26 minutes. Assuming linear acceleration, this means $47\cdot(22\cdot3600 + 26\cdot60)/1566=2424$ seconds on average per airfoil (using a single process). Accordingly, PSO takes about $1.65\cdot 10^5$ times more time than DRL to optimise each airfoil using a single process. This is because, once trained, the DRL agent does not need to make any solver call to optimise an airfoil. Since training is a one-time cost, DRL becomes increasingly advantageous when optimising multiple airfoils, as its inference is virtually instantaneous. 
This highlights the drastic efficiency advantage of DRL, making PSO impractically slow by comparison. 


\subsection{Hyperparameter exploration in the hybrid optimisation case}\label{sec:multi}
\vspace{-5mm}
Here we provide additional insights on the performance of the DRL agent trained on the hybrid optimisation case (i.e. $\sigma >0$ in Eq.~\ref{eq:reward_with_condifence_definitive}). We systematically vary $\sigma$ --where larger values of $\sigma$ provide more weight to the structural optimisation over the aerodynamic one), and assess
(i) the number of training steps needed to reach training convergence, (ii) the lift-to-drag ratio performance metrics \texttt{improvement} and \texttt{best} evaluated on all the airfoils present in the Aerosandbox library and (iii) the variation in the maximum thickness, as a function of the regulariser's hyperparameter $\sigma$. 
Results (for the converged airfoils) are depicted in Table~\ref{tab:keep_area_results}, which serves as a sensitivity analysis of $\sigma$. It can be observed that the number of necessary training steps varies (typically increases) with $\sigma$. 
At the same time, results confirm the evidence illustrated in Fig~\ref{fig:comparison_penalization_area} and show that the DRL agent (i) is able to improve the aerodynamic efficiency while (ii) also maintaining the structural integrity of the airfoil. Finally, results also certify that forcing the maximum thickness to be preserved comes at the cost of a smaller improvement of the aerodynamic efficiency.


\begin{table}[h]
    \centering
    \begin{tabular}{|r|c|c|c|c|r|}
        \hline
        \textbf{$\sigma$} & \textbf{Improvement} & \textbf{Best} & {$\Delta\textsc{mt}$} {(\%)}& \textbf{\shortstack{Airfoils \\ Evaluated}} & \textbf{\shortstack{Training \\ steps}} \\
        \hline
        \hline
        0   & $140\pm49$  & 
        241 (37) & $64\pm22$  & 1982  & 81920   \\
        2   & $108\pm41$  & 
        201 (49) & $22\pm16$  & 1965  & 120832  \\
        5   & $99\pm39$   & 
        190 (52) & $21\pm18$  & 1960  & 120832  \\
        10  & $95\pm42$   & 
        190 (53) & $15\pm10$  & 1909  & 120832  \\
        15  & $91\pm37$   & 
        180 (46) & $12\pm11$  & 1952  & 120832  \\
        20  & $93\pm37$   & 
        186 (45) & $12\pm10$  & 1933  & 120832  \\
        30  & $94\pm41$   & 
        186 (57) & $8\pm11$   & 1938  & 501760  \\
        100 & $84\pm42$   & 
        176 (59) & $5\pm5$    & 1922  & 1001470 \\
        1000& $54\pm35$   & 
        137 (46) & $7\pm8$    & 1897  & 501760  \\
        \hline
    \end{tabular}
    \caption{Hybrid optimisation agent evaluation with UIUC for different values of $\sigma$. Pointwise statistics of the \texttt{Improvement} and \texttt{best} metrics are provided as in Table \ref{tab:optimisation_comparison_uiuc}, and $\Delta \textsc{mt}$ statistics over the whole UIUC is provided in terms of mean $\pm$ one standard deviation.}
    \label{tab:keep_area_results}
\end{table}

\subsection{Enhancing the DRL-based airfoil optimisation with Transfer Learning}
\vspace{-5mm}
To round off, here we compare the training efficiency and evaluate the performance of the trained DRL agent when training is done using a single aerodynamic solver
({\it XFoil} alone or, for completeness, {\it NeuralFoil} alone) with respect to the case when the agent is pre-trained with {\it NeuralFoil} and fine-tuned with {\it XFoil}). For the sake of exposition, we separate the purely aerodynamic optimisation ($\sigma=0$ in Eq.~\ref{eq:reward_with_condifence_definitive}) and the hybrid one ($\sigma>0$).

\medskip \noindent 
{\bf Purely-aerodynamic optimisation case.} Results for all DRL agents in the purely aerodynamic case (with and without transfer learning) are summarised in Table~\ref{tab:neuralfoil_fine-tuning_evaluation}, reporting (i) the total number of training steps, (ii) the total solver time, (iii) the number of airfoils evaluated, and the metrics (iv) pointwise statistics of \texttt{best} and (v) \texttt{improvement} over all the airfoils in the evaluation set. Pointwise statistics of \texttt{best} reported in the table are given in terms of median (IQR) as the distributions are not Gaussian (Shapiro-Wilk test) and highly skewed, whereas in the case of \texttt{improvement} we report mean $\pm$ std as the distributions are fairly more symmetrical.
Note that to compute the total solver time we simply multiply the number of training steps by the execution time per call to the particular aerodynamic solver, which according to \textit{NeuralFoil}'s GitHub amounts to $73$ ms for a call to {\it XFoil} and $4$ ms for a call to the smallest version of {\it NeuralFoil}, which is the one we actually used. Although the total solver time is not equivalent to the total training time, it is nonetheless an objective way of comparing the computational effort of the agent using vs. not using transfer learning.
As previously mentioned, the number of airfoils evaluated may differ slightly between agents because {\it XFoil}, which is consistently used for evaluation, occasionally fails to converge. In such cases, we excluded the corresponding airfoils from the analysis.
\begin{table*}[htb]
\caption{\label{tab:neuralfoil_fine-tuning_evaluation}%
Comparison of training efficiency and evaluation performance of DRL agents: one trained from scratch using \textit{XFoil}, one pre-trained with \textit{NeuralFoil}, and two using transfer learning (TL). Solver time is estimated from 73\,ms/call for \textit{XFoil} and 4\,ms/call for \textit{NeuralFoil}. Performance metrics are evaluated on the UIUC dataset.}
\begin{ruledtabular}
\begin{tabular}{lccccc}
DRL Agent & 
\shortstack{Training steps} & 
\shortstack{Solver time (s)} & 
\shortstack{Airfoils evaluated} & 
Improvement & 
Best \\
\hline
Fully trained with \textit{XFoil} (no TL) & 81920 & 5980 & 1982 & $140 \pm 49$ & 241(37) \\
Pre-training (\textit{NeuralFoil}) & 26312 & 105 & 1970 & $66 \pm 32$ & 154(33) \\
Transfer Learning (strategy \#3) & 10240 & 748 & 1983 & $120 \pm 43$ & 219(29) \\
Transfer Learning (strategy \#1) & 10240 & 748 & 1977 & $136 \pm 44$ & 236(28) \\
\end{tabular}
\end{ruledtabular}
\end{table*}

\medskip \noindent
First, out of the four transfer learning strategies depicted in Sec.~\ref{sec:transfer_learning}, we found that strategy $\# 1$ 
had the best performance, followed by $\# 3$. Strategies $\# 2$ and $\# 4$ did not converge and are therefore not reported in the table. This lack of convergence indicates that the information present in the final layers of the neural network plays an important role in shaping the agent’s behaviour.

\medskip \noindent
Second, from the table we observe that the number of necessary training steps for the agent solely trained with {\it Xfoil} (81920) is notably larger than the total number of steps needed for the DRL agents with transfer learning, where the pre-training ({\it NeuralFoil}) and fine-tuning totally amounts to $26312+10240=36552$ training steps. The cost of the steps performed with {\it NeuralFoil} could be considered negligible, so the important aspect here is that the number of steps required with {\it Xfoil} is reduced by a factor of 8. Additionally, it can be observed that the total number of iterations decreases. This is because the agent converges with fewer iterations when trained with {\it NeuralFoil}, something that can be attributed to its continuous nature and its lack of non-convergence problems. This is already evidence that transfer learning reduces the computational effort. 

\medskip \noindent
Third, not only the number of steps, but also the total solver time during training ($5980$ seconds vs. $105+748=853$ seconds) is notably shorter for agents trained using transfer learning. On this respect, we have also compared the total training time with and without transfer learning. If we define the \texttt{Time Reduction} ($\%$) as

\begin{widetext}
\begin{equation}
\texttt{Time Reduction (\%)} = 
\frac{
    \text{TL-free training time} - \text{TL-enhanced training time}
}{
    \text{TL-free training time}
} \cdot 100\;. \label{eq:reduction}
\end{equation}
\end{widetext}

we find $\texttt{Time Reduction} \approx 86 \%$, reaching higher reductions than in other settings \cite{BHOLA2023112018}.

\begin{figure}[htb!]
    \centering
\includegraphics[width=0.8\linewidth]{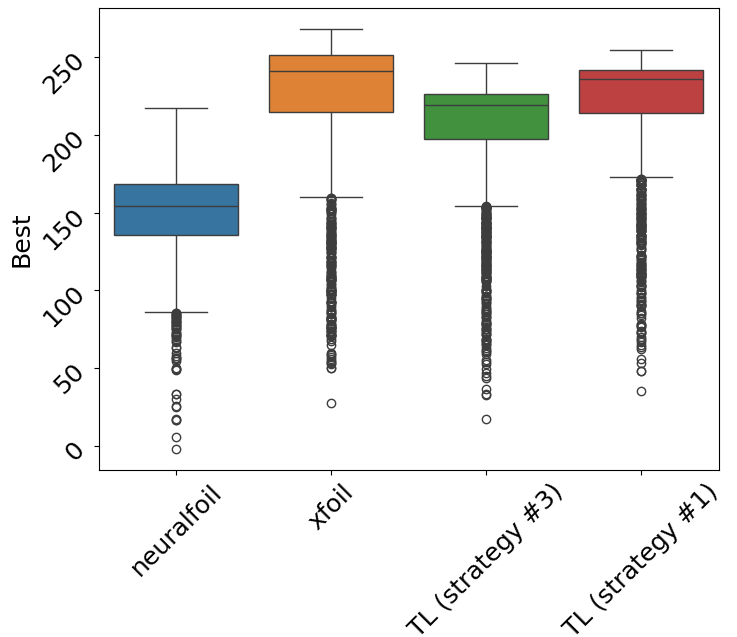}
    \caption{Boxplot (showing medians, interquantile range (IQR) and whiskers at $\pm 1.5$ IQR)) of the \texttt{best} performance metric in the evaluation set, for all (purely-aerodynamic) DRL agents. Points outside whiskers are considered outliers, these emerge as the underlying distribution is left-skewed. The DRL agent trained solely on {\it Xfoil} remains statistically superior (Mann-Whitney U test), but the performance obtained by the DRL agent trained with transfer learning's strategy $\#1$ is very close.}
    \label{fig:boxplots_3_1}
\end{figure}

\medskip \noindent
Fourth, the performance in the evaluation set of the agent trained solely on {\it XFoil} seems statistically superior (Mann-Whitney U test $p< 10^{-9}$), but the DRL agent trained with Transfer Learning (strategy $\# 1$) shows a very similar performance (paired sign test $p\approx 0.006$, see boxplots of the \texttt{best} performance metric in Fig.~\ref{fig:boxplots_3_1}), confirming that transfer learning indeed increases computational efficiency while maintaining good performance.

\begin{figure}[htb!]
    \centering
\includegraphics[width=0.48\linewidth]{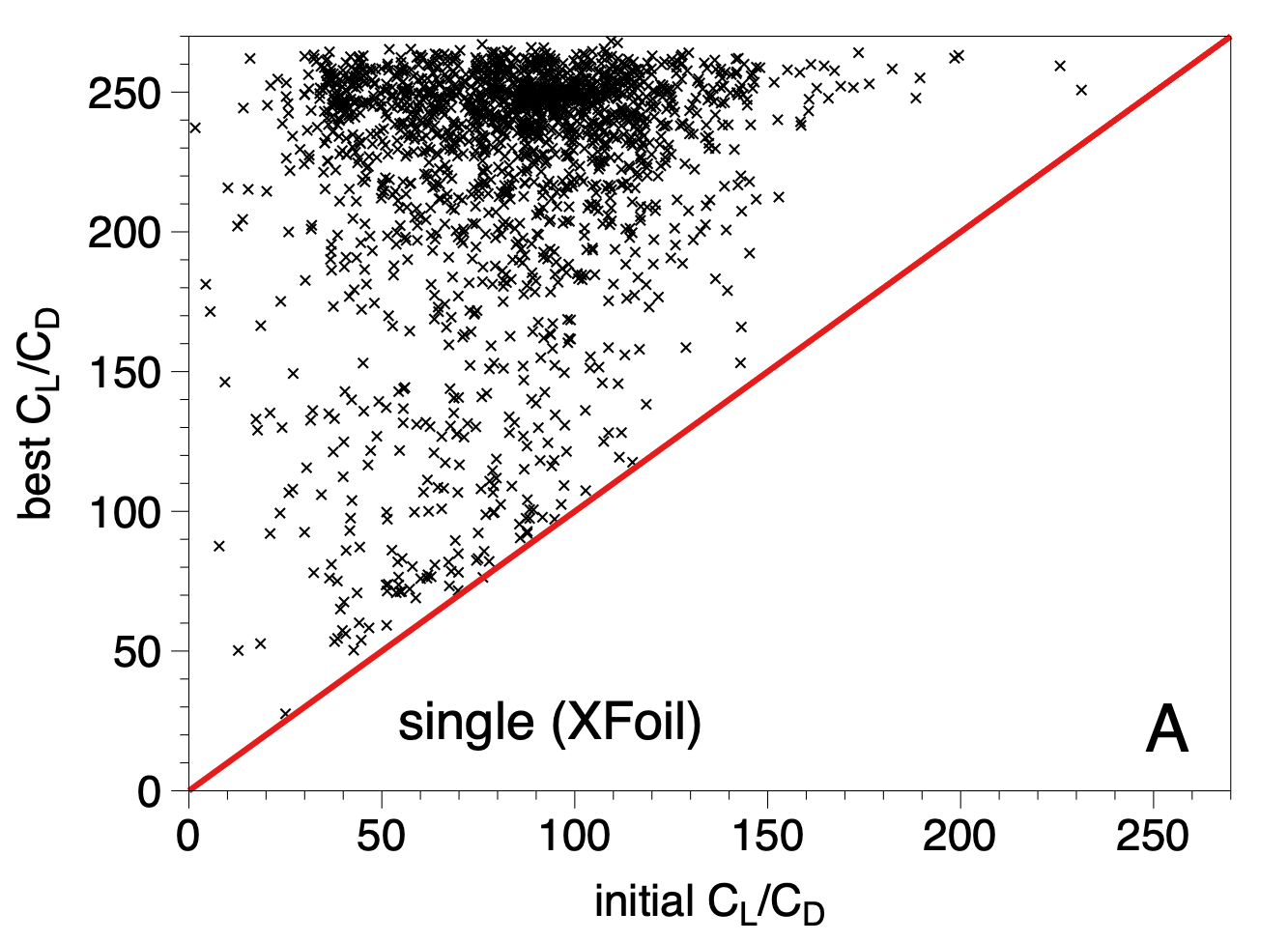}
\includegraphics[width=0.48\linewidth]{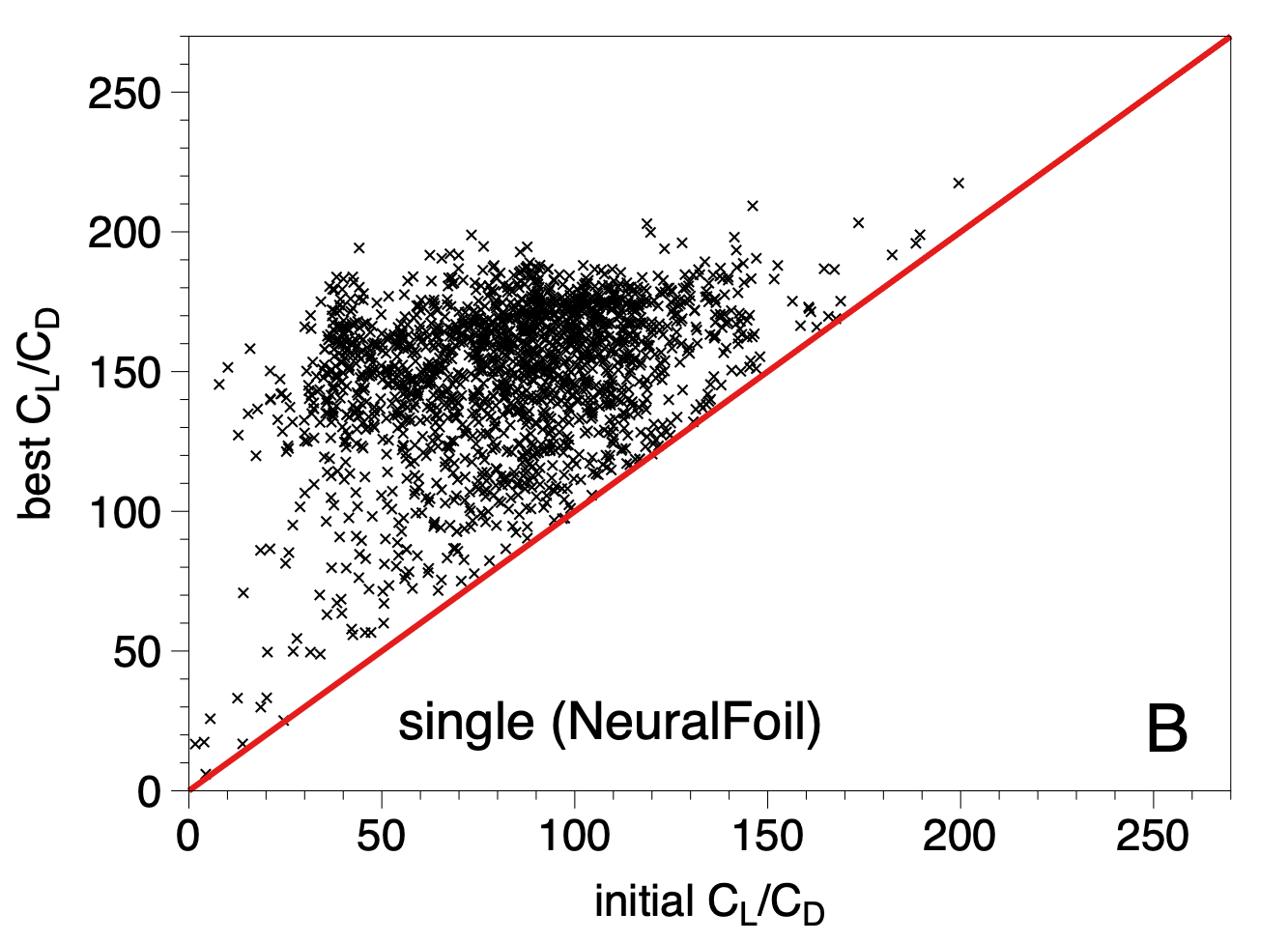}
\includegraphics[width=0.48\linewidth]{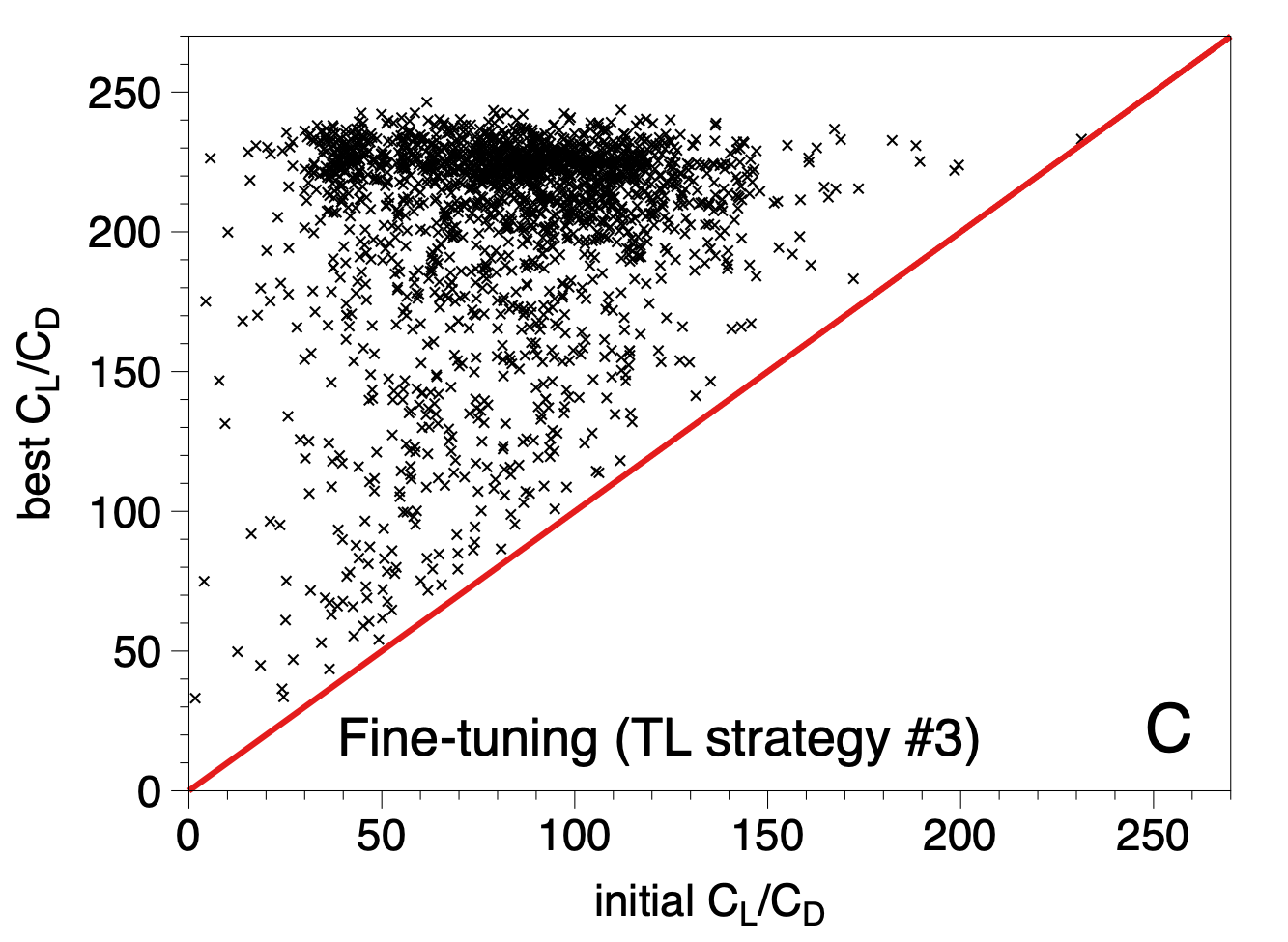}
\includegraphics[width=0.48\linewidth]{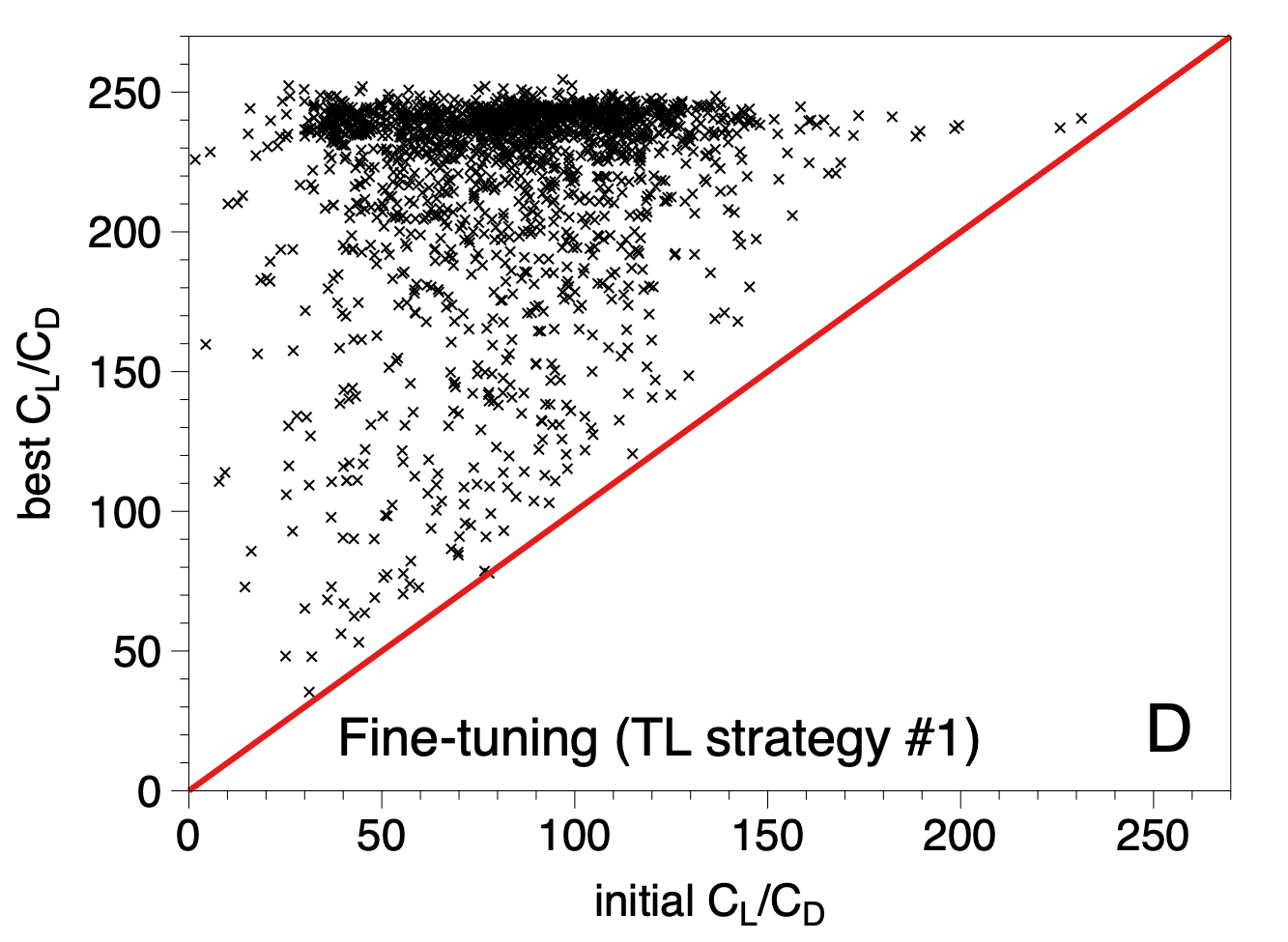}
    \caption{Scatter plots of the metric \texttt{best} as a function of the initial lift-to-drag ratio for all airfoils in the evaluation set, as computed from (A) the DRL agent trained with {\it XFoil}, (B) the pre-training with {\it NeuralFoil}, (C) the DRL agent trained with transfer learning (strategy $\# 3$), and (D) the DRL agent trained with transfer learning (strategy $\# 1$), within the purely-aerodynamic optimisation scenario.}
    \label{fig:scatters}
\end{figure}

\medskip \noindent
Finally, in Fig.~\ref{fig:scatters} we scatter plot the best lift-to-drag ratio obtained for each evaluated airfoil as a function of the initial estimate of the airfoil shape for the four DRL agents. These scatter plots show that while there are some specific shapes for which the agent cannot improve the aerodynamic efficiency, in most of the cases all DRL agents can improve it, with varying degrees of performance, which are in good agreement with those previously reported. Interestingly, we find that all airfoils with initial $C_L/C_D > 150$ seem to be greatly improved by the DRL agent trained solely with {\it XFoil} and the one trained with the Transfer Learning strategy $\# 1$ (see the clear exclusion area in panels (A) and (D)).

\medskip \noindent 
{\bf Hybrid optimisation case.} Results of the DRL agent (with TL) trained on a combined aerodynamic/structural optimisation 
{are depicted in Table~\ref{tab:TL_max_thickness}. Comparing these with the equivalent results without TL reported in Table~\ref{tab:keep_area_results},  
we find that (i) the procedure involving TL substantially reduces the number of training steps as well as the training time, reaching  $\texttt{Time Reduction}$ as large as $94\%$;
(ii) fixing the structural hyperparameter $\sigma$, the improvement in aerodynamic efficiency is only slightly lower for every value of $\sigma$ (around a 10\%) in the TL-enhanced DRL agent than the one without TL, and (iii) the preservation of the maximum thickness is quite similar in both techniques. In summary, TL-enhancement also works in the hybrid optimisation case.

\begin{table*}[htb]
\caption{\label{tab:TL_max_thickness}%
Evaluation of the TL-enhanced hybrid aerodynamic/structural DRL agent on the UIUC dataset, for different values of the structural hyperparameter $\sigma$. Statistics for \texttt{Improvement}, \texttt{Best}, and $\Delta \textsc{mt}$ are given pointwise. \texttt{Time Reduction} is computed via Eq.~\ref{eq:reduction} by comparing training time with and without TL.}
\begin{ruledtabular}
\begin{tabular}{ccccccc}
$\sigma$ & Improvement & Best & $\Delta \textsc{mt}$ (\%) & \shortstack{Training steps} & \shortstack{Airfoils evaluated} & \shortstack{Time reduction (\%)} \\
\hline
2     & $98 \pm 37$  & 189 (46) & $20 \pm 16$ & 25088  & 1945 & 78.0 \\
5     & $92 \pm 34$  & 182 (30) & $15 \pm 15$ & 25088  & 1961 & 78.0 \\
10    & $87 \pm 36$  & 178 (38) & $13 \pm 14$ & 25088  & 1950 & 78.0 \\
15    & $85 \pm 36$  & 171 (40) & $16 \pm 16$ & 35328  & 1952 & 69.7 \\
20    & $82 \pm 36$  & 171 (41) & $14 \pm 13$ & 35328  & 1958 & 69.7 \\
30    & $79 \pm 36$  & 171 (46) & $14 \pm 13$ & 35328  & 1958 & 92.6 \\
100   & $62 \pm 35$  & 151 (45) & $8 \pm 8$   & 50176  & 1905 & 94.8 \\
1000  & $26 \pm 22$  & 110 (41) & $4 \pm 5$   & 50176  & 1802 & 89.7 \\
\end{tabular}
\end{ruledtabular}
\end{table*}

\section{Conclusions}
\label{sec:conclusions}
\vspace{-3mm}
While traditional airfoil shape optimisation typically considers aerodynamic efficiency, the optimisation can produce airfoils whose structural integrity might be compromised; in other words, there is a need to preserve some structural constraints --such as the airfoil's maximum thickness-- within the optimisation process.
In this paper, we have developed a technique for airfoil shape optimisation --flexible to accomodate purely aerodynamic or hybrid (aerodynamic/structural) shape optimisation-- that enhances a deep reinforcement learning (DRL) agent with Transfer Learning (TL).   
Our results indicate that (i) DRL works both for the aerodynamic and aerodynamic/structural shape optimisation problems, (ii) DRL is notably superior to other gradient-free methods such as Particle Swarm Optimisation (PSO) in terms of computational efficiency (being several orders of magnitude faster) and aerodynamic performance, although PSO seem to enforce thickness constraints almost exactly, while DRL does that only approximately, (iii) and the TL-enhancement reduces further training costs with respect to the TL-free case around a 86\%, while approximately maintaining performance (in terms of aerodynamic efficiency and structural preservation).


\begin{figure}[htb!]
    \centering
\includegraphics[width=1\linewidth]{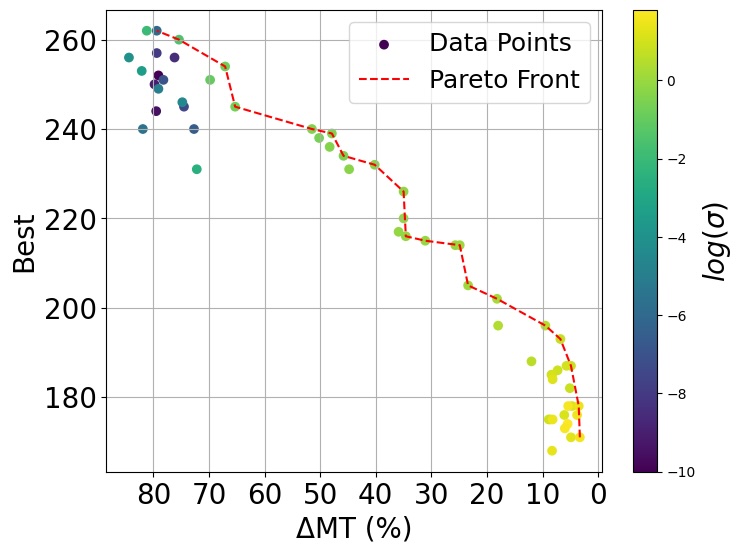}
    \caption{Scatter plot of the DRL-based hybrid optimisation, showing the aerodynamic improvement metric $\texttt{Best}$ as a function of the structural integrity preservation metric $\Delta \textsc{mt}$, for a variety of hyperparameters $\sigma$. The Pareto front is highlighted, although note that our initial multi-objective optimisation problem is scalarised and therefore Pareto optimality is not guaranteed.}
    \label{fig:Pareto}
\end{figure}

\medskip \noindent In summary, our results suggest that DRL stands as an efficient method for airfoil optimisation, both in terms of increasing aerodynamic efficiency and also in the search of geometries that at the same time have enhanced aerodynamic properties and realistically preserve structural constraints. 

\medskip \noindent 
On relation to the use of TL-enhanced DRL, our results encourage the exploration of more challenging problems, such as e.g. the optimisation of (3-dimensional) wings: a problem where both the computational cost of the solvers and the complexity of the shape parameterisation increase significantly,  and where transfer learning techniques that use inexpensive surrogate solvers could be of great help.
More generally, the efficiency of this optimisation technique might enable the incorporation of gold standard -yet expensive- aerodynamic solvers, such as CFD-based optimisation, something that also deserves future investigation. 
Finally, a more general and exciting open problem is to 
assess whether the benefits of this combined technique also apply in other problems that encompass aerodynamic and structural shape optimisation, e.g. maximisisation of lift-to-drag ratio or flow,
minimisation of mass, pressure drop, or stress concentration, and maintaining strength or stiffness, and their application to e.g. pipe shapes, car frames, or fillet optimisation to cite a few. Other interesting applications where DRL could be of great use include the design of propellers, rotor blades, and turbine components in aerospace; aerodynamic shaping of vehicles and structural frames in automotive engineering; and flow optimisation in energy systems like HVAC ducts, wind farms, and hydraulic networks.

\medskip \noindent
On relation to the aerodynamic vs structural optimisation scenario, in this work we initially chose to scalarize the initial multi-objective optimisation problem by combining the aerodynamic and the structural integrity objectives into a single reward function. For illustration, we advance in Fig.~\ref{fig:Pareto} the results of a Pareto front analysis, where the values of both aerodynamic and structural integrity improvements are plotted for a variety of values of the hyperparameter $\sigma$. Further work is needed to  consider both aerodynamic and structural terms independently as well as to add more sophisticated structural criteria, in order to embrace full-fledged multi-objective optimisation.



\section*{Supplementary Material}
\vspace{-5mm}
Supplementary material contains URLs where animations of the optimisation process of the airfoils in table \ref{tab:airfoil_comparison_pso_rl} are available for the different algorithms described.

\section*{Acknowledgments}
\vspace{-5mm}
\noindent The authors thank useful discussions within the NUMATH group. DR, EV and GR  acknowledge funding from project TIFON (PLEC2023-010251) funded by MCIN/AEI/10.13039/501100011033, Spain, funding from the European Union (project HERFUSE) under GA No 101140567. 
Views and opinions expressed are those of the authors only, and do not necessarily reflect those of the European Union or Clean Aviation Joint Undertaking. Neither the European Union nor Clean Aviation JU can be held responsible for them.
LL acknowledges partial financial support from via project MISLAND (PID2020-114324GB-C22), CSxAI (PID2024-157526NB-I00) and the Maria de Maeztu Program for units of Excellence (Grant No. CEX2021-001164 M), the three of them funded by MICIU/AEI/10.13039/501100011033, and funding from the European Commission Chips Joint Undertaking project No. 101194363 (NEHIL).
GR acknowledges partial financial support received by the Grant DeepCFD (Project No. PID2022-137899OB-I00) funded by MICIU/AEI/10.13039/501100011033 and by ERDF, EU. 
Finally, all authors gratefully acknowledge the Universidad Politécnica de Madrid (www.upm.es) for providing computing resources on Magerit Supercomputer.

\bibliography{bibliography.bib}

\appendix

\section{RL environment configuration}
\label{ap:env_configuration}
\vspace{-5mm}
This section shows the configuration used for the environments. The ranges of the CST parameters, shown in Table \ref{tab:env_configuration_cst},  were selected based on empirical testing, aiming to strike a balance between allowing the agent sufficient freedom to explore the design space and avoiding excessively extreme or unrealistic geometries.

\begin{table*}[!htbp]
\caption{\label{tab:env_configuration_cst}%
CST parameter bounds used in the environments. Each surface is represented by 8 control points.}
\begin{ruledtabular}
\begin{tabular}{lcc}
\textbf{Parameter} & \textbf{Lower bound} & \textbf{Upper bound} \\
\hline
Upper surface & $[-1.5, -1.5, -1.5, -1.5, -1.5, -1.5, -1.5, -1.5]$ & $[1.25, 1.25, 1.25, 1.25, 1.25, 1.25, 1.25, 1.25]$ \\
Lower surface & $[-0.75, -0.75, -0.75, -0.75, -0.75, -0.75, -0.75, -0.75]$  & $[1.5, 1.5, 1.5, 1.5, 1.5, 1.5, 1.5, 1.5]$ \\
Trailing-edge thickness & $0.0005$ & $0.01$ \\
Leading-edge weight & $-0.05$ & $0.775$ \\
\end{tabular}
\end{ruledtabular}
\end{table*}

\noindent The initial airfoil for an episode is selected from this list of 20 NACA airfoils:
naca0006, naca0009, naca0012, naca0015, naca0018, naca1408, naca1410, naca1412, naca2412, naca2415, naca4412, naca4415, naca4420, naca6412, naca6415, naca7421, naca8409, naca8412, naca8415, naca9421.

\section{PPO hyperparameters}
\label{ap:ppo_params}
\vspace{-5mm}
This section provides a summary of the PPO hyperparameters used to train the different agents discussed in this work.
Sample inefficiency is a common problem for many reinforcement learning algorithms, which means that to learn, the agent needs to see too many simulations. For problems like this, where the simulations are expensive, this could be an issue. To minimise the number of required simulations and consequently lower the computational cost of the training process, the PPO hyperparameters have been adjusted with this objective in mind. After many trainings where these hyperparameters were being modified, it has been observed, as expected, that the most relevant parameter in terms of the number of steps needed to train an agent is the discount factor, $\gamma$, and the second one, the clip range. Moreover, it was found empirically that the optimum value of $\gamma$ is 0.3, and the value of clip range was increased, to a greater or lesser extent, from its default value 0.2. For instance, with this hyperparameter tuning and using Neuralfoil, an agent can be trained on 60000 steps that has the same performance as one trained on 120000 steps using the default hyperparameters given by Stable Baselines 3. Since our primary goal was to reduce training cost through transfer learning, we did not perform an exhaustive hyperparameter search and considered the empirical tuning sufficiently effective for our purposes. Nonetheless, we conducted a hyperparameter search using \textit{Optuna} \cite{optuna_2019} for the \textit{NeuralFoil} agent. It is worth noting that the resulting configuration closely aligns with the parameters previously obtained through empirical tuning.
Figure \ref{fig:comparacion_gammas} shows the mean reward that the agent obtained during training for different values of $\gamma$. As can be observed, the line for $\gamma=0.3$ grows faster than the line for $\gamma=0.99$, which means that the agent trained with $\gamma=0.3$ is learning with fewer steps.

\begin{figure}[!htbp]
    \centering
    \includegraphics[width=1\linewidth]{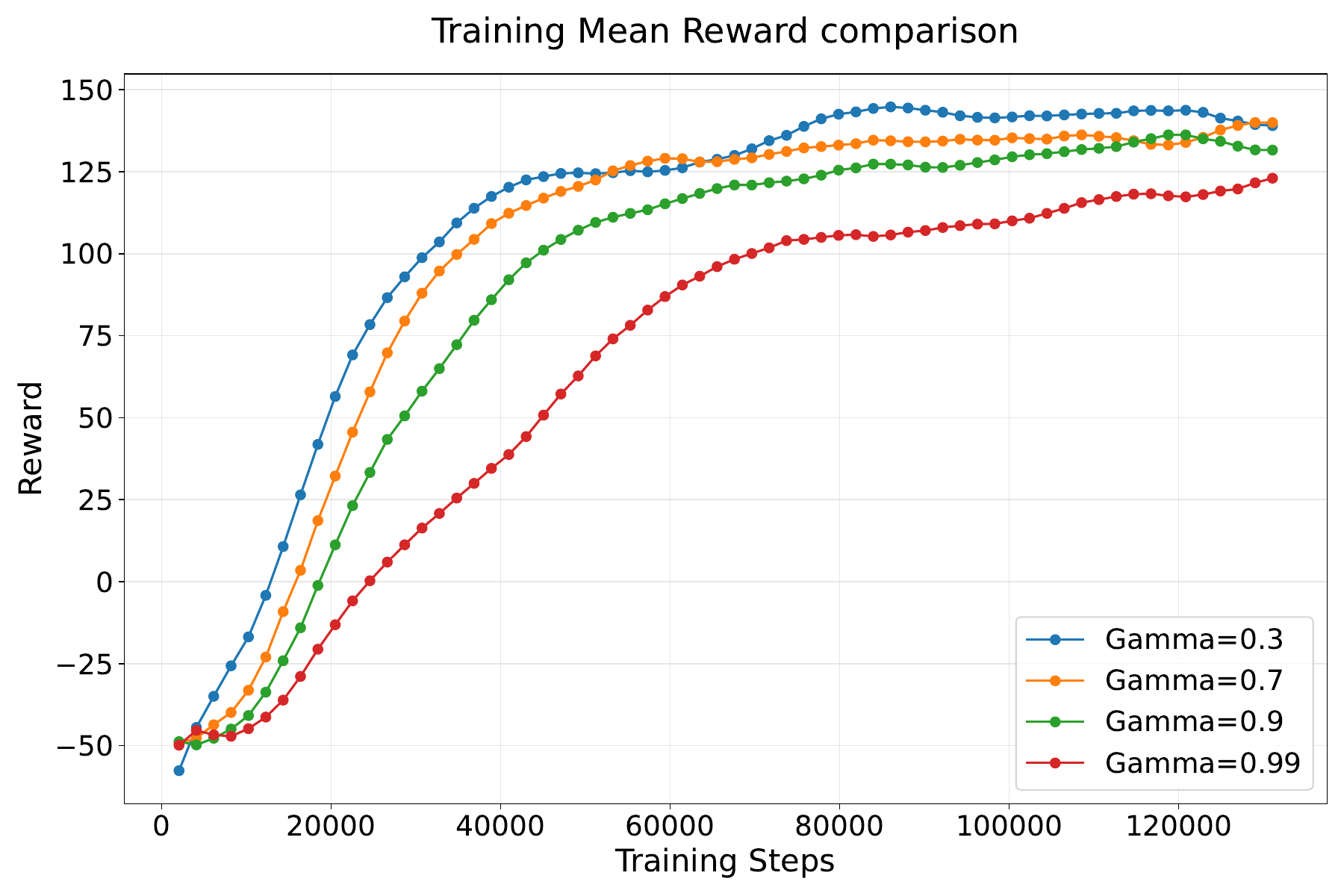}
    \caption{Episode mean reward during training with different discount factors ($\gamma$)}
    \label{fig:comparacion_gammas}
\end{figure}

\noindent 
Table~\ref{tab:PPO_hyperparameters_all} below consolidates the PPO hyperparameters for three training configurations: one agent trained from scratch using \textit{XFoil}, the pre-trained agent with \textit{NeuralFoil} and the fine-tuned from a previously trained \textit{XFoil} policy.
If a parameter is not included in the table, the default value given by Stable Baselines 3 will be used.

\begin{table}[!htbp]
    \centering
    \begin{tabular}{l|c|c|c}
        \hline
        \textbf{Solver} & \textbf{\textit{XFoil}} & \textbf{\textit{NeuralFoil}} & \textbf{\textit{XFoil}} \\
        \textbf{TL strategy} & \textbf{from scratch} & \textbf{pre-training } & \textbf{transfer learning} \\        
        \hline
        Total timesteps & - & 26312 & 10240 \\
        Learning rate & $2.5 \times 10^{-4}$ & $2.5 \times 10^{-4}$ & $2.5 \times 10^{-4}$ \\
        Steps per update & 2048 & 2048 & 512 \\
        Batch size & 64 & 64 & 64 \\
        Num. epochs & 20 & 10 & 20 \\
        Discount factor ($\gamma$) & 0.3 & 0.3 & 0.3 \\
        Clip range & 0.3 & 0.6 & 0.2 \\
        Entropy coefficient & 0.001 & 0 & 0.005 \\
        Policy net arch. & [256, 256] & [256, 256] & [256, 256] \\
        Value net arch. & [256, 256] & [256, 256] & [256, 256] \\
        \hline
    \end{tabular}
    \caption{PPO hyperparameters used for training the different agents presented in Table~\ref{tab:neuralfoil_fine-tuning_evaluation} of the main text. The total number of timesteps for the agent trained from scratch with \textit{XFoil} varies and is specified in its corresponding section}
    \label{tab:PPO_hyperparameters_all}
\end{table}

\section{\textit{XFoil} configuration}
\label{ap:xfoil_configuration}
\vspace{-5mm}
This section details the configuration settings used for \textit{XFoil} during the generation of simulation data for training and evaluation. The selected parameters, shown in Table~\ref{tab:xfoil_config}, were chosen to ensure a balance between computational efficiency and numerical stability.

\begin{table}[!htbp]
    \centering
    \begin{tabular}{ll}
        \hline
        \textbf{Parameter} & \textbf{Value} \\
        \hline
        Maximum Iterations & 200 \\
        Number of Panels & 255 \\
        Critical Amplification Factor ($n_{crit}$) & 9 \\
        Timeout & 30 seconds \\
        Angle of Attack ($\alpha$) & 2° \\
        Reynolds Number (Re) & $1 \times 10^6$ \\
        Mach Number (Ma) & 0.5 \\
        \hline
    \end{tabular}
    \caption{Configuration parameters used for \textit{XFoil} simulations. These settings define the aerodynamic conditions and solver parameters employed during training and evaluation}
    \label{tab:xfoil_config}
\end{table}

\nocite{*}

\end{document}